\documentclass[preprint,12pt]{elsarticle}

\usepackage[page]{appendix}

\usepackage{graphics} % for pdf, bitmapped graphics files
\usepackage{epsfig} % for postscript graphics files
\usepackage[outdir=./]{epstopdf}
\usepackage{mathptmx} % assumes new font selection scheme installed
\usepackage{times} % assumes new font selection scheme installed
\usepackage{amsmath} % assumes amsmath package installed
\usepackage{amssymb}  % assumes amsmath package installed
\usepackage{bm} 
\usepackage[x11names,table,xcdraw]{xcolor}
\usepackage{graphicx}
\usepackage{comment}
\usepackage{makecell}
\usepackage{enumitem}
\usepackage[T1]{fontenc}
\usepackage{mathtools}
\usepackage{marginnote}
\usepackage{tgbonum}
\usepackage{tabularx}
\usepackage{multirow}
\usepackage{subcaption}
\usepackage{algorithm}
\usepackage{algpseudocode}
\usepackage{threeparttable}
\usepackage{booktabs}
\usepackage{dcolumn}
\usepackage{diagbox}
\newcolumntype{d}[1]{D{.}{.}{#1}}

\algrenewcommand\algorithmicindent{0.5em}%

\usepackage{newtxtext, newtxmath}

\usepackage[pagebackref=true]{hyperref} 
\renewcommand*\backref[1]{\ifx#1\relax \else (Cited~on~p.~#1) \fi}

\usepackage{sepfootnotes}

\usepackage{xspace}

\newtheorem{theorem}{Theorem}

\newtheorem{proposition}[theorem]{Proposition}

\journal{Control Engineering Practice}

\newcommand{\Kinova}{\textsc{Kinova}\xspace}

\newcommand{\Matlab}{\textsc{MATLAB}\xspace}
\newcommand{\Simscape}{\textsc{Simscape Multibody}\xspace}

\newcommand{\eigenval}{\ensuremath{\lambda}\xspace}
\newcommand{\SameElemMat}{\ensuremath{*}\xspace}

\newcommand{\ContIdx}{\ensuremath{\text{c}}\xspace}

\newcommand{\StateVec}{\ensuremath{\mathbf{x}}\xspace}

\newcommand{\InputVec}{\ensuremath{\mathbf{u}}\xspace}

\newcommand{\StateMat}{\ensuremath{\mathbf{A}^{\ContIdx}}\xspace}
\newcommand{\StateMatDisc}{\ensuremath{\mathbf{A}}\xspace}
\newcommand{\InputMat}{\ensuremath{\mathbf{B}^{\ContIdx}}\xspace}

\newcommand{\NumSchedVar}{\ensuremath{n_{\phi}}\xspace}
\newcommand{\SchedVec}{\ensuremath{\Phi}\xspace}

\newcommand{\SchedVar}{\ensuremath{\phi}\xspace}
\newcommand{\SchedVarIdx}{\ensuremath{\SchedVar_i}\xspace}
\newcommand{\ConvexHull}{\ensuremath{\bm{\Theta}}\xspace}
\newcommand{\NumVertex}{\ensuremath{n_{v}}\xspace}
\newcommand{\IdxVertex}{\ensuremath{i}\xspace}
\newcommand{\HullVertex}{\ensuremath{\bm{\theta}}\xspace}
\newcommand{\HullVertexIdx}{\ensuremath{\HullVertex_\IdxVertex}\xspace}

\newcommand{\LyapMatrix}{\ensuremath{\mathbf{P}}\xspace}
\newcommand{\GainAnalysisMatrix}{\ensuremath{\mathbf{L}}\xspace}
\newcommand{\InvLyapMatrix}{\ensuremath{\mathbf{Y}}\xspace}
\newcommand{\LyapCand}{\ensuremath{V}\xspace}
\newcommand{\SelectMatrix}{\ensuremath{\mathbf{S}}\xspace}

\newcommand{\IdentMat}{\ensuremath{\mathbf{I}}\xspace}
\newcommand{\ZeroMatrix}{\ensuremath{\mathbf{0}}\xspace}

\newcommand{\Ellipsoid}{\ensuremath{\mathcal{E}}\xspace}
\newcommand{\VarEllipsoid}{\ensuremath{\mathbf{z}}\xspace}

\newcommand{\Dstab}{\ensuremath{\mathbb{D}}-stability\xspace}

\newcommand{\LDstab}{\ensuremath{\bm{\alpha}}\xspace}
\newcommand{\MDstab}{\ensuremath{\bm{\beta}}\xspace}

\newcommand{\DampingRatio}{\ensuremath{\mathbf{\xi}}\xspace}
\newcommand{\PercOverShoot}{\ensuremath{OS}\xspace}
\newcommand{\LogSpiral}{\ensuremath{\varphi}\xspace}
\newcommand{\IntersectOS}{\ensuremath{a_0}\xspace}
\newcommand{\CenterOS}{\ensuremath{a_{se}}\xspace}
\newcommand{\MajorAxOS}{\ensuremath{a_e}\xspace}
\newcommand{\MinorAxOS}{\ensuremath{b_e}\xspace}
\newcommand{\ConeAngle}{\ensuremath{\gamma}\xspace}

\newcommand{\VICSol}{\ensuremath{\mathcal{C}_{s}}\xspace}
\newcommand{\VICSolScore}{\ensuremath{f_{s}}\xspace}

\newcommand{\InertVIC}{\ensuremath{H}\xspace}
\newcommand{\DampVIC}{\ensuremath{D}\xspace}
\newcommand{\StiffVIC}{\ensuremath{K}\xspace}
\newcommand{\PosError}{\ensuremath{e}\xspace}
\newcommand{\VelError}{\ensuremath{\dot{\PosError}}\xspace}
\newcommand{\AccError}{\ensuremath{\ddot{\PosError}}\xspace}

\newcommand{\gainVIC}{\ensuremath{\mathbf{W}^{\ContIdx}}\xspace}
\newcommand{\gainVICDisc}{\ensuremath{\mathbf{W}}\xspace}
\newcommand{\ForceVIC}{\ensuremath{F}\xspace}
\newcommand{\InputMatForceVIC}{\ensuremath{\InputMat_{\ForceVIC}}\xspace}

\begin{document}

\begin{frontmatter}



\title{Condition-based Design of Variable Impedance Controllers from User Demonstrations}


\author{Alberto San-Miguel\corref{mycorrespondingauthor}}
\cortext[mycorrespondingauthor]{Corresponding author}
\ead{asanmiguel@iri.upc.edu}

\author{Vicen\c{c} Puig}
\author{Guillem Aleny\`{a}}

\address{Institut de Rob\`{o}tica i Inform\`{a}tica Industrial CSIC-UPC,\\ Llorens i Artigas, 4-6, 08028 Barcelona, Spain}

\begin{abstract}
	
This paper presents an approach to ensure conditions on Variable Impedance Controllers through the off-line tuning of the parameters involved in its description. In particular, we prove its application to term modulations defined by a Learning from Demonstration technique. This is performed through the assessment of conditions regarding safety and performance, which encompass heuristics and constraints in the form of Linear Matrix Inequalities. Latter ones allow to define a convex optimisation problem to analyse their fulfilment, and require a polytopic description of the VIC, in this case, obtained from its formulation as a discrete-time Linear Parameter Varying system. With respect to the current state-of-art, this approach only limits the term definition obtained by the Learning from Demonstration technique to be continuous and function of exogenous signals, i.e. external variables to the robot. Therefore, using a solution-search method, the most suitable set of parameters according to assessment criteria can be obtained. Using a 7-DoF \Kinova \textsc{Gen3} manipulator, validation and comparison against solutions with relaxed conditions are performed. The method is applied to generate Variable Impedance Controllers for a pulley belt \textit{looping} task, inspired by the Assembly Challenge for Industrial Robotics in World Robot Summit 2018, to reduce the exerted force with respect to a standard (constant) Impedance Controller. Additionally, method agility is evaluated on the generation of controllers for one-off modifications of the nominal belt \textit{looping} task setup without new demonstrations.

\end{abstract}

%\begin{figure*}[tbp]
%	\centerline{\includegraphics[width=0.5\linewidth]{./Figures/Initial_scheme.png}}
%	\captionsetup{labelformat=empty}
%	\caption{From human-guided demonstrations, Learning from Demonstrations techniques provide the reference trajectory to follow together with the required compliance profile for the task. The method presented in this paper (LMI-based VIC Design) takes this to derive the set of parameters that define a Variable Impedance Controller (VIC), considering constraints on safety (stability and operational limits) and performance (considering User Preference mechanism). The person appearing in this Figure is the first author and gives permission to use his image for this purpose.}
%\label{fig:method_scheme}
%\end{figure*}

\begin{comment}
\begin{highlights}
\item Ensuring operation conditions on robot controllers is paramount in anthropic domains.
\item For LfD-based VICs, only stability has been addressed by limiting term modulations.
\item Presented method ensures more conditions for LfD-based VICs by tuning its parameters.
\item Conditions regard safety and performance, in terms of LMI constraints and heuristics.
\item The most suitable set of parameters is found off-line through an automated process.
\item The method is validated and applied to a real case study using a \textsc{KINOVA} manipulator.
\end{highlights}
\end{comment}


\end{frontmatter}

\section{Introduction} \label{sec:intro}

Great research efforts are devoted to introduce robots in anthropic domains (both industrial and domestic) for the sake of further enhancing tasks by physically interacting with humans and the environment. This calls for techniques that, firstly, formalise task characteristics as structures such that, secondly, can be used by control strategies for execution. For the first part, Learning from Demonstrations (LfD) techniques allow the generation of task descriptions through multiple human-guided demonstrations, from which relevant information can be extracted~\cite{ravichandar2020recent}. This approach is suitable for users that are not familiar with robotic platforms and does not require an iterative execution process until a successful solution is found as with Reinforcement Learning (RL) based techniques~\cite{kober2013reinforcement}. For the second part of the problem, Impedance Control (IC) schemes have arisen as a trade-off between classical position and force tracking control, such that the relationship between them is tracked instead~\cite{NevilleHogan1985}. This approach is especially relevant in those scenarios where the robot must follow a trajectory but physical interactions (with humans or the environment) might happen or are even necessary for task completion. Moreover, many tasks require or benefit from the modulation of impedance terms throughout the task, namely Variable Impedance Control (VIC)~\cite{ikeura1995variable}. Therefore, there has been a recent interest in integrated solutions that make use of the LfD paradigm to generate modulation profiles for VIC terms~\cite{Abu-Dakka2020}. 

However, in this intersection of techniques, it is required to ensure the fulfilment of some conditions that determine a reliable execution of the task. In all application contexts, ensuring stability is the paramount concern. For VIC, on top of requiring positive terms as for IC, term modulations need to be considered for stability assessment. Many approaches in the literature make use of methods based on Lyapunov theory, which proves stability through the existence of a suitable candidate function~\cite{Behal2009}. This has led to the derivation of sufficient conditions on modulation profiles (joint or individual) as in~\cite{Kronander2016}. In this line, advances have been made in simplifying them through the imposition of particular structures for modulation terms, e.g. with an online filter as in~\cite{Bednarczyk2020}. Other approaches consider energy-based strategies, e.g. through the so-called energy tanks~\cite{Ferraguti2013} that ``store'' all the dissipative effects (energy-wise) for performing non-dissipative movements, which might be ill-posed by its dependency on robot state and initialization. Within LfD context, the approach presented in~\cite{Khansari2015a} defines stability conditions (for pre-defined forms of term modulations) and discusses its compatibility with the use of learning techniques, which was shown afterwards in~\cite{Khader2021a}. Besides stability, up to our knowledge, only a few works addressing standard IC consider the introduction of other guarantees, e.g.~\cite{bednarczyk2020model}, but without addressing its joint use with LfD techniques.

\begin{figure}[tbp]
    \centerline{\includegraphics[width=\linewidth]{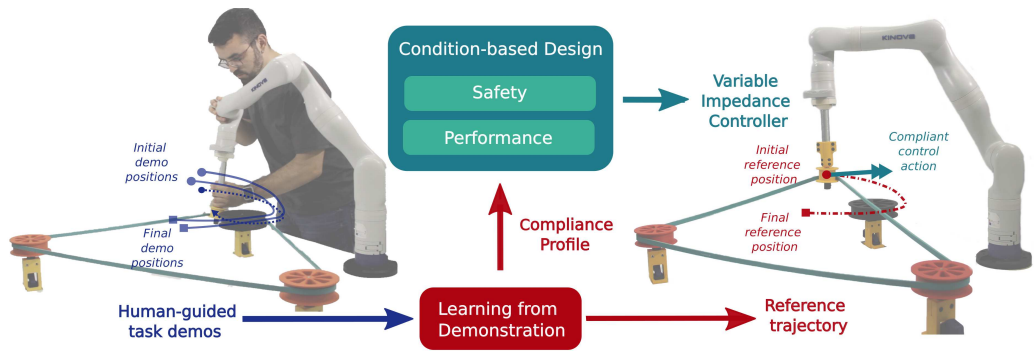}}
    \caption{\footnotesize From human-guided demonstrations, LfD is used to provide the reference trajectory to follow together with the required compliance profile for the task. This is used to define a Variable Impedance Controller, and the proposed approach (Condition-based Design) provides the set of parameters that complete its definition such that safety and performance conditions are fulfilled. The person appearing in this Figure is the first author and gave permission to use his image for this purpose.}
    \label{fig:LfDVIC_method_scheme}
\end{figure}

In this paper, we contribute with a systematic framework to simultaneously guarantee multiple conditions on VICs by offline tuning its parameters. Particularly, we put the focus on those obtained from a LfD technique to extract a compliance profile, expanding our previous work~\cite{san2022automated}. Conditions can limit an operation region for the controller, i.e. aimed at safety (e.g. stability), or determine how its operation suits the desired behaviour for the task, i.e. aimed at performance. For those defined as constraints, VIC operation is embedded by a polytopic description generated through its formulation as a Linear Parameter Varying (LPV) system, which w.r.t. state-of-art approaches, only limits modulations to be continuous and function of exogenous signals to the robot. Moreover, this description is given for its discrete-time form to acknowledge controller implementation in the real robot, which is executed in a fixed-frequency control loop. Conditions are arranged into an assessment problem such that the most suitable set of parameters that makes VIC fulfil them is generated. Figure~\ref{fig:LfDVIC_method_scheme} conceptualises the proposed approach for the chosen LfD technique in the context of the case study presented in this work, consisting of a pulley belt \textit{looping} task to generate VIC that reduce required force w.r.t. constant IC.

The paper is organised as follows: Sect.~\ref{sec:LfDVIC_ProblemStatement} details the VIC description used to exemplify the method, based on the well-known LfD technique presented in~\cite{Calinon2010}. Section~\ref{sec:LfDVIC_PolytopicForm} details the procedure to obtain the polytopic description for the VIC from LPV formulation. The assessment process is described in Sec.~\ref{sec:LfDVIC_SolAssess} including chosen conditions on safety and performance. Section~\ref{sec:LfDVIC_AutomatedSol} portraits the complete method integrating the VIC description generation from demonstrations and the iterative assessment together with the solution-search method. Design validation experiments have been included in Sec.~\ref{sec:LfDVIC_Validation} and its application for the case study in Sec.~\ref{sec:LfDVIC_case_study}. Finally, conclusions on the method are drawn in Sec.~\ref{sec:LfDVIC_Conclusions} together with a discussion on the results and future work. 
\section{Problem Statement} \label{sec:LfDVIC_ProblemStatement}
For the sake of simplicity, let's consider a 1-DoF task. Impedance Control (IC) aims at imposing a second-order dynamic relationship between (external) force $F(t)$ and system motion. For a trajectory tracking task, it is represented by the position error $e \coloneqq p^{r}(t) - p(t)$ and its derivatives, w.r.t. a reference trajectory $\{p^r, \dot{p}^r, \ddot{p}^r\}^{T}_{t=0}$. Thus, the IC relationship is characterised by a set of terms that leverage motion variables, namely inertia $H$ (for acceleration), damping $D$ (for velocity) and stiffness $K$ (for position). As aforementioned, Variable Impedance Controllers (VICs) vary these terms to change robot behaviour throughout a task. In this paper, the focus is put on a VIC with time-variant stiffness
\begin{equation} \label{eq:LfDVIC_imp_relation}
	H\,\AccError(t) + D \, \VelError(t) + K(t) \, \PosError(t) = \ForceVIC(t)
	\vspace{-1mm} \\.
\end{equation}
In this work, LfD is used to generate the modulation profile for $K(t)$. Particularly, through the well-known technique presented in~\cite{Calinon2010}, which links variance in position over different demonstrations for the same trajectory to the required compliance. Thus, lower variability, i.e. small position differences at the same time, will require lower compliance, i.e. high stiffness. For this purpose, a Heteroscedastic Gaussian Process (H-GP) model is considered to embed all the demonstrations and extract the required data. This model consists of two standard GP models: one on the mean of the observations and another on the variance between observations, i.e. $\mathcal{N}(\bm{\mu}(t),\bm{\Sigma}(t))$. Thus, H-GP can be generated through Expectation-Maximization (EM) on the data set $\mathcal{D} =\{\{(t_{m,n},p_{m,n}^d)\}^{T_n}_{n=1}\}^{M}_{m=1}$ conformed of a set of $M$ position demonstrations $p^d_n$ aligned for times $t_n$ of size $T_m$, as presented in~\cite{kersting2007most}. From this description, reference trajectory is described by the mean, i.e. $\mu(t) = p^r(t)$, and $\dot{p}^r$ and $\ddot{p}^r$  obtained through differentiation, and eigenvalues $\lambda(t)$ of the inverse covariance matrix $\bm{\Sigma}^{-1}(t)$ are used to define the variant stiffness $K$ along time as
\begin{equation} \label{eq:LfDVIC_K_scaling}
	K(t) = K_{min} + (K_{max} - K_{min})\frac{\text{log}(\lambda(t))-\text{log}(\lambda_{min})}{\text{log}(\lambda_{max})-\text{log}(\lambda_{min})} \\.
\end{equation}
Considering this, controller solution $\mathcal{C}_s$ can be defined as the tuple of parameters that unequivocally determine the behaviour of~\eqref{eq:LfDVIC_imp_relation}. Assuming $H$ is given beforehand:
\begin{equation}\label{eq:LfDVIC_Sol_Definition}
	\mathcal{C}_s = \langle K_{max}^s,\;K_{min}^s, \; D \rangle.
\end{equation}
Hence, the method presented in this paper aims at finding the most suitable $\mathcal{C}_s$ such that the VIC fulfils a set of conditions regarding safety and performance. 

At this point, it is important to note that, although a 1~-~DoF controller is presented, the method is not limited to a particular number nor a certain combination of DoFs. Also, it should be mentioned that we make use of this particular LfD technique to showcase our approach with a well-understood compliance generation paradigm. Instead, the proposed approach can be applied to any other technique that generates modulation profiles (function of continuous exogenous signals) and simultaneously modulates any of the parameters. These topics will be further discussed in Sec.~\ref{sec:LfDVIC_Conclusions}.

%\newpage
\section{Polytopic description for VIC} \label{sec:LfDVIC_PolytopicForm}

Some of the conditions imposed for the VIC require to consider its complete operation range, i.e. include its term modulations in their assessment. In this work, the chosen approach consists of embedding its operation range through a polytopic description defined by its limits. 

\subsection{State-space formulation}

Considering $\StateVec(t) \coloneqq [\PosError(t)\;\VelError(t)]^T$, impedance relationship~\eqref{eq:LfDVIC_imp_relation} can be stated into continuous-time state-space form:
\begin{equation}\label{eq:LfDVIC_imp_state_space}
	\dot{\StateVec}(t) =\StateMat(t)\cdot\StateVec(t)+\InputMatForceVIC\cdot \ForceVIC(t), 
\end{equation}
being $\mathbf{B}_F$ the force input matrix and $\StateMat$ the state matrix, defined as
\begin{equation} \label{eq:LfDVIC_imp_state_space_matrix_def}
	\StateMat(t) = 
	\begin{bmatrix} 
		0 & 1 \\  
		-\StiffVIC(t)\cdot\InertVIC^{-1} & -\DampVIC\cdot\InertVIC^{-1} 
	\end{bmatrix}\,,\qquad
	\InputMat_F = 
	\begin{bmatrix} 
		0 \\ 
		\InertVIC^{-1}
	\end{bmatrix}\,.
\end{equation}
State matrix \StateMat can be further divided into a constant term $\StateMat_{0}$ and the control effort $\InputVec(t)$ corresponding to the VIC :
\begin{equation} \label{eq:LfDVIC_imp_state_mat_def}
	\StateMat(t) = \StateMat_{0}\cdot\StateVec(t) + \InputMat\cdot \InputVec(t)
\end{equation}
being
\begin{equation*} \label{eq:ForceVIC_imp_ss_state_mat_def_terms}
	\StateMat_{0} = 
	\begin{bmatrix} 
		0 & 1 \\  
		0 & 0 \\ 
	\end{bmatrix}\,,\qquad
	\InputMat = 
	\begin{bmatrix} 
		0 \\ 
            1
	\end{bmatrix}\,, \qquad
	\InputVec(t) = \textbf{W}(t)\cdot\StateVec(t)\,;
\end{equation*}
where the variant VIC gain is:
\begin{equation*} \label{eq:ForceVIC_imp_ss_VIC_gain}
	\textbf{W}(t) = 
	\begin{bmatrix} 
		-\StiffVIC(t)\cdot\InertVIC^{-1} & -\DampVIC\cdot\InertVIC^{-1}
	\end{bmatrix}\,.
\end{equation*}
The nature of impedance relationship~\eqref{eq:LfDVIC_imp_relation} renders~\eqref{eq:LfDVIC_imp_state_space} as a linear relationship with time dependency terms, namely Linear Time Variant (LTV).

%MIGHT BE INTERESTING TO SAY THAT IT IS GENERALIZABLE TO THE MULTI AXIS CASE
%Regard that this description is given for the case where \InertVIC, \DampVIC and \StiffVIC  full-matrices. Generally, in the case of \adhoc VIC descriptions, modulation laws  imposed separately for each DoF, in this case, translational axis in the Cartesian Space. This means that there  no couplings between matrices and, therefore, these matrices  diagonal. Hence, in the context of this work, each impedance relationship can be designed separately. This is the case for the handover task scenario, where only the control over a single DoF is considered. To avoid misinterpretations, the notation of these matrices remains disregarding their dimensions. 

\subsection{LPV Model} \label{sec:LfDVIC_Formulation_PolytopicLPV}
If the variant terms of a LTV system can be arranged into a set of varying parameters $\SchedVec = [\SchedVar_1,...,\SchedVarIdx,... ,\SchedVar_{\NumSchedVar}]$, the system is referred as Linear Parameter Varying (LPV)~\cite{shamma2012overview}. In this work, the LPV form of~\eqref{eq:LfDVIC_imp_state_space} is obtained through the embedding of non-linear terms into \SchedVarIdx, following~\cite{Kwiatkowski2006}. This approach constrains each scheduling variable (i) to be a-priori known, measured or estimated on-line and (ii) continuous and defined in the operation range. Hence:
\begin{equation} \label{eq:LfDVIC_sched_parameter_def}
	\SchedVec(t) \equiv \SchedVar_1 (t) = - \StiffVIC(t)\cdot\InertVIC^{-1}.    
\end{equation}
such that the LPV formulation is
\begin{equation} \label{eq:LfDVIC_LPV_state_space}
	\dot{\StateVec}(t) = \StateMat(\SchedVec(t)) \cdot \StateVec(t) + \InputMatForceVIC\cdot \ForceVIC(t),
\end{equation}
being
\begin{equation} \label{eq:LfDVIC_LPV_def}
	\StateMat(\SchedVec) = 
\begin{bmatrix} 
	0 & 1 \\  
	\SchedVar_1(t) & -\DampVIC\cdot\InertVIC^{-1}
\end{bmatrix}\,,\qquad 
\textbf{W}(\SchedVec) = 
\begin{bmatrix} 
-\SchedVar_1(t)& -\DampVIC \cdot\InertVIC^{-1}
\end{bmatrix}\,.
\end{equation}
Considering a non-zero \InertVIC, variant stiffness $K(t)$ is (i) off-line determined from demonstrations given $K_{max}$ and $K_{min}$ through~\eqref{eq:LfDVIC_K_scaling} and (ii) being defined from a linear operation over the covariance matrix of the H-GP model, is continuous and defined throughout time.

\begin{figure}[tbp]
	\centerline{\includegraphics[width=\linewidth]{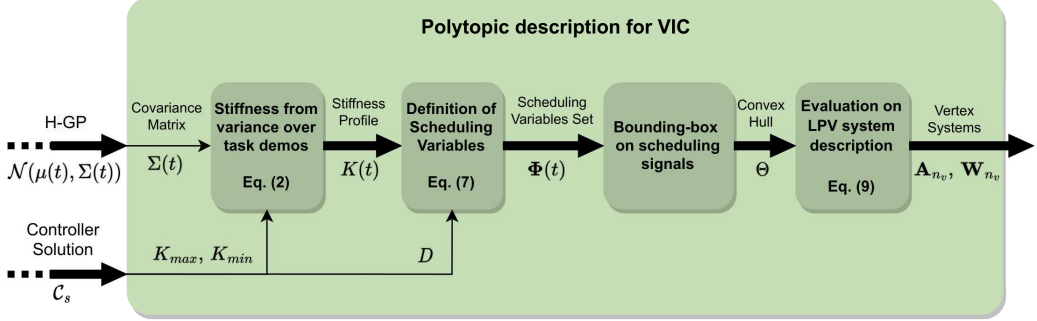}}
	\caption{Generation of the polytopic description of the VIC from controller solution and the H-GP model embedding task demonstrations.}
	\label{fig:LfDVIC_Generation_Polytope}
\end{figure}

\subsection{Polytopic representation} 

Properties can be assessed for LPV systems, but dealing with all their (infinite) reachable operating points defined by $\SchedVec(t)$ is not numerically tractable. Instead, using a polytopic description of~\eqref{eq:LfDVIC_LPV_state_space} allows to deal only with a set of Linear Time Invariant (LTI) systems at the limits of operation, namely vertex systems. First, all the trajectories of $\SchedVec(t)$ have to be enclosed within the convex hull $\ConvexHull$ with \NumVertex vertices $\HullVertexIdx$:
\begin{equation} \label{eq:convex_hull}
	\SchedVec(t) \in \ConvexHull \coloneqq \text{Co}\{\HullVertex_1,...,\HullVertexIdx,...,\HullVertex_{\NumVertex}\}.
\end{equation}
In this work the so-called \textit{bounding box} method~\cite{sun1998affine} is applied such that each vertex is the combination of lower and upper bounds of each varying parameter $\underline{\SchedVarIdx}$ and $\overline{\SchedVarIdx}$, which leads to $\NumVertex = 2 ^{\NumSchedVar}$. For the VIC case, these limits can be obtained through the evaluation of~\eqref{eq:LfDVIC_sched_parameter_def} throughout the extracted stiffness profile from~\eqref{eq:LfDVIC_K_scaling} for the complete execution. Finally, aforementioned set of vertex systems that conform the polytopic representation can be obtained by evaluating the LPV model at each vertex of $\bm{\Theta}$. Using the LPV definitons~\eqref{eq:LfDVIC_LPV_def}:
\begin{equation} \label{eq:LfDVIC_VertexSystems}
	[\StateMat_{i}\,,\;\gainVIC_{i}] \coloneqq  [\StateMat(\HullVertex_{i})\,,\;\gainVIC(\HullVertex_{i})] 
\end{equation}
Using this formulation, each $\mathcal{C}_s$ can be linked to a set of vertex systems to be used in the forthcoming assessment of conditions. Figure~\ref{fig:LfDVIC_Generation_Polytope} summarises the complete process in the context of this work. It is important to note that
definitions~\eqref{eq:LfDVIC_VertexSystems} are for the continuous-time form of the system. As aforementioned, to acknowledge the real execution on robotic platforms, controller assessment conditions are stated for discrete implementations. In this work, according to~\cite{toth2010discretisation}, their discrete-time equivalents are be obtained by evaluating system's matrices according to a discretisation method for a sampling time $T_s$. Details on the particular application to the platform used in work are included in Sect.~\ref{sec:LfDVIC_Validation}.

\section{VIC Solution Assessment}\label{sec:LfDVIC_SolAssess}

The method proposed in this work aims at finding controller solutions \VICSol that define a VIC fulfilling a set of conditions. Hence, they are stated into an assessment problem of \VICSol, differentiating between those regarding safety and performance. First ones are understood as those that bound controller operation by limiting a region within the \VICSol space. On the other hand, those on performance define criteria that determine VIC suitability for the task within the \VICSol space. Thus, the complete assessment is used to find solutions at the intersection of both, i.e. those suitable for the task with a bounded operation. Moreover, some of these conditions can be formulated as constraints over the complete operation range of the VIC thanks to its polytopic representation. Thus, these constraints can be stated in the form of Linear Matrix Inequalities (LMIs) to each of the vertex system, such that, if a common solution is found, the constraint is fulfilled for the complete operation range. With respect to our previous work~\cite{san2022automated}, additional safety conditions in the form of LMIs are included and performance one corresponds to an heuristic that introduces user intuition over the required compliant behaviour. 

\subsection{Safety}

As aforementioned, the paramount concern when designing a controller is to preserve stability, which can be formulated as a constraint in the form of LMIs. 

\begin{proposition} \label{prop:LfDVIC_Stab} 
	\textit{Stability LMI Constraints for VIC.} Considering the discrete form of the polytopic description~\eqref{eq:LfDVIC_VertexSystems} for LPV model~\eqref{eq:LfDVIC_LPV_state_space}, the equilibrium $\StateVec = 0$ is stable in the sense of Lyapunov for $k = [0,\infty) \in \mathbb{N} $ if there exist a solution matrix $\LyapMatrix>\mathbf{0}\,|\,\LyapMatrix = \LyapMatrix ^T$ that simultaneously fulfils the following LMI $\forall i = 1,...,\NumVertex$:
	\begin{equation}\label{eq:LfDVIC_LMI_Stab}
		\StateMatDisc_i^{T}\cdot \LyapMatrix \cdot \StateMatDisc_i - \LyapMatrix \leq \ZeroMatrix
	\end{equation}
\end{proposition}
\noindent \textit{Proof:} The proof is given in App.~\ref{appendix:LfDVIC_LMI_Proof_Stab}.

\vspace{5mm}

Besides stability, it is desirable to set limits over variables that define controller behaviour. This
allows to set guarantees over task execution beforehand. For VIC, upper bounds for position error and control effort can be simultaneously introduced using LMI constraints.

\begin{proposition} \label{prop:LfDVIC_limits}
	\textit{Maximum Effort and Error LMI Constraints for VIC.} Considering the discrete form of the polytopic description~\eqref{eq:LfDVIC_VertexSystems} for LPV model~\eqref{eq:LfDVIC_LPV_state_space}, conditions
	\begin{subequations} \label{eq:LfDVIC_limits}
		\begin{gather}
			\InputVec(k)  \, \leq \, \InputVec _{max}\;, \qquad 
			|e_k|  \, \leq \, \Delta p_{max} \; ; \tag{\theequation a,b}
		\end{gather}
	\end{subequations}
	 satisfied for $k=[0,\infty)\in\mathbb{N}$ and an initial state $\StateVec(0)$ if there exist a solution matrix $\LyapMatrix>\mathbf{0}\,|\,\LyapMatrix = \LyapMatrix ^T$ that simultaneously fulfils the following LMI $\forall i = 1,...,\NumVertex$,
	\begin{subequations}\label{eq:LfDVIC_LMI_Operational}
		\begin{gather}
			\begin{bmatrix} %\label{eq:LfDVIC_LMI_control_limit}
				\InputVec_{max}^2 \cdot \IdentMat_{\gainVICDisc} & \gainVICDisc_{i}\\
				\gainVICDisc_{i} ^T & \LyapMatrix
			\end{bmatrix}\geq \ZeroMatrix \; , \qquad 
			\begin{bmatrix} %\label{eq:LfDVIC_LMI_error_limit}
				\LyapMatrix & \StateMatDisc_i^T \cdot \mathbf{S}^T \\
				\SelectMatrix \cdot \StateMatDisc_i &  \Delta p_{max}^2 \\
			\end{bmatrix}\geq \ZeroMatrix \,;   \tag{\theequation a,b}
		\end{gather}
	\end{subequations}
	where $\IdentMat_{\gainVICDisc}$ is a matrix of the appropriate dimensions and $\mathbf{S}$ a selection matrix for $e(t)$; and such that 
	\begin{equation}\label{eq:LfDVIC_LMI_init_state}
		\begin{bmatrix} 
			\IdentMat_0 & {\StateVec(0)}^T \cdot \LyapMatrix \\
			\LyapMatrix \cdot \StateVec(0)  & \LyapMatrix
		\end{bmatrix}\geq \ZeroMatrix \,;
	\end{equation}
	being $\IdentMat_{0}$ a matrix of the appropriate dimensions.
\end{proposition}
\noindent \textit{Proof:} The proof is given in App.~\ref{appendix:LfDVIC_LMI_Proof_limits}.

\vspace{5mm}
Another relevant characteristic defining control behaviour is the transient response of the system, i.e. how it behaves until reaching the steady state. In second-order systems like \eqref{eq:LfDVIC_imp_relation} exponentially decaying oscillations might appear. This phenomenon is associated to damping ratio $\DampingRatio \in (0,1)$, i.e. underdamped systems. This can be characterised through the Percentage Overshooting (\PercOverShoot), which corresponds to the maximum peak value of the system measured from the reference, and can be expressed as function of $\DampingRatio$. Hence, in this work, vertex systems' poles are confined into a region in the discrete complex plane which imposes a maximum damping ratio $\overline{\DampingRatio}$ that corresponds to a maximum Percentage Overshooting $\overline{\PercOverShoot}$. This is defined through LMI constraints based on the concept of \Dstab.

%\begin{comment}
\begin{proposition} \label{prop:LfDVIC_Overshoot}
	\textit{Maximum Overshooting LMI Constraints for VIC.}  Considering the discrete form of the polytopic description~\eqref{eq:LfDVIC_VertexSystems} for LPV model~\eqref{eq:LfDVIC_LPV_state_space}, system's response will not surpass the maximum percentage overshoot  $\overline{\PercOverShoot}$ if there exist a solution matrix $\LyapMatrix>\mathbf{0}\,|\,\LyapMatrix = \LyapMatrix ^T$ that simultaneously fulfils the following LMI $\forall i = 1,...,\NumVertex$:
	\begin{equation}\label{eq:LfDVIC_LMI_Dstab_OS}
		\LDstab \otimes \LyapMatrix + \MDstab \otimes (\LyapMatrix \cdot \StateMatDisc_i) + \MDstab^T \otimes (\StateMatDisc_i^T \cdot \LyapMatrix) \leq \ZeroMatrix
	\end{equation}
	being $\LDstab = \text{diag}(\LDstab_e,\;\LDstab_v)$, $\MDstab = \text{diag}(\MDstab_e,\;\MDstab_v)$ defined according to \cite{Rosinova2019} as follows:
	\begin{subequations}\label{eq:LfDVIC_Dstab_OS_def}
		\begin{align}
			\LDstab_e &= \begin{bmatrix} %\label{eq:LfDVIC_def_alpha_e}
				-1 & -\frac{a_{se}}{a_e}\\
				\SameElemMat & -1
			\end{bmatrix}\,, \; &\MDstab_e = \frac{1}{2} \begin{bmatrix} %\label{eq:LfDVIC_def_beta_e}
				0 & \frac{1}{a_e}-\frac{1}{b_e} \\
				\frac{1}{a_e}+\frac{1}{b_e} & 0
			\end{bmatrix} \,; \tag{\theequation a,b} \\	
			\LDstab_v &= -2 \cdot \begin{bmatrix} %\label{eq:LfDVIC_def_alpha_v}
				\text{sin}(\gamma) & 0\\
				\SameElemMat & \text{sin}(\gamma) \end{bmatrix}\,, \;
			&\MDstab_v = \begin{bmatrix} %\label{eq:LfDVIC_def_beta_v}
				\text{sin}(\gamma) & \text{cos}(\gamma) \\
				-\text{cos}(\gamma) & \text{sin}(\gamma)
			\end{bmatrix} \,; \tag{\theequation c,d}
		\end{align}
	\end{subequations}
	where
	\begin{subequations}\label{eq:LfDVIC_Dstab_OS_def_terms}
		\begin{gather}
			\overline{\DampingRatio}  = -\text{log}(\overline{\PercOverShoot}/100)/\sqrt{\smash[b]{\pi^2 + \text{log}^2(\overline{\PercOverShoot}/100)}}\, ; \tag{\theequation a}\\
			\LogSpiral = \text{cos}^{-1}(\overline{\DampingRatio})\,, \qquad \IntersectOS = -e^{-\pi/\text{tan}{\varphi}}\,; \tag{\theequation b,c}\\
			\qquad \CenterOS = (1+\IntersectOS)/2 \qquad \MajorAxOS= (1-\IntersectOS)/2 \;; \tag{\theequation d,e}\\
			\MinorAxOS = b \cdot \MajorAxOS/\sqrt{\smash[b]{\MajorAxOS^2 - (a-\CenterOS)^2}}\; ; \tag{\theequation f}\\
			\ConeAngle = \text{tan}^{-1}(b/(1-a)). \tag{\theequation g}
		\end{gather}
	\end{subequations}
	where $\text{Re}(\mathbf{r}) = a$ and $\text{Im}(\mathbf{r}) = b$, being $\mathbf{r}$ in the complex discrete plane belonging to the logarithmic spiral defined by \LogSpiral.
\end{proposition}
\noindent \textit{Proof:} The proof is given in App.~\ref{appendix:LfDVIC_LMI_Proof_Overshoot}.
%\end{comment}

\subsection{Performance}

\begin{figure}[!tbp]
	\centerline{\includegraphics[width=\linewidth]{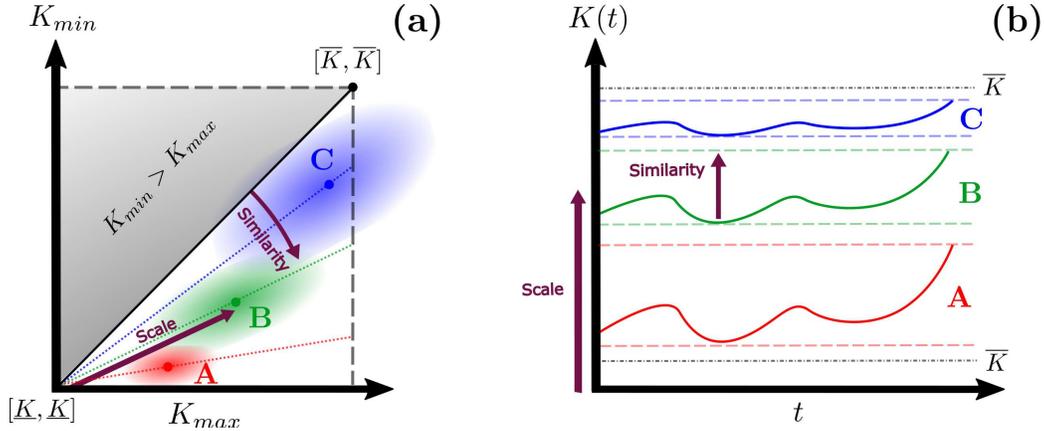}}
	\caption{Graphical representation of the User Preference mechanism for three different cases on the same compliance profile in $\StiffVIC_{max}-\StiffVIC_{min}$ plane (a) and the corresponding \StiffVIC profiles (b). Case A has both low Scale and Similarity values, which translates into a stiffness modulation $\StiffVIC(t)$ with a $\StiffVIC_{min}$ closer to $\underline{\StiffVIC}$ and very distinct from $\StiffVIC_{max}$. For B, both $\StiffVIC_{min}$ and $\StiffVIC_{max}$ have higher values than in A as the Scale is higher, but also a higher Similarity makes their values closer. Case C has both high Scale and Similarity values and therefore both $\StiffVIC_{min}$ and $\StiffVIC_{max}$ are closer to $\overline{\StiffVIC}$ with similar values. }
	\label{fig:user_pref}
\end{figure}

How successfully a task is performed by a robot, in many cases, depends on how well human intuition is introduced into its execution. The LfD technique used in this work only allows to specify the compliance profile through a set of demonstrations on the same task, but the user has no means to introduce the desired rigidity associated with each maximum and minimum level of compliance. Hence, in this work, performance is assessed through an heuristic that characterises desired compliance, namely User Preference. Particularly, this mechanism is defined in the controller solution plane region described by the maximum $\overline{\StiffVIC}$ and minimum $\underline{\StiffVIC}$ values of $\StiffVIC_{max}$ and $\StiffVIC_{min}$ through two parameters ranging in $[0,1]$: Similarity and Scale. Similarity defines the ``closeness" between $_{max}$ and $\StiffVIC_{min}$, and Scale where the profile remains within $[\underline{\StiffVIC},\overline{\StiffVIC}]$. Mathematically, this is represented by a Gaussian that maps the tuple $(\StiffVIC_{max},\StiffVIC_{min})$ to a value in $[0,1]$, being the $0$ value at its centre. Similarity determines the angle between the diagonal $\StiffVIC_{max} = \StiffVIC_{min}$ and the direction of the major axis of the Gaussian function passing by $(\underline{\StiffVIC},\underline{\StiffVIC})$. Its centre is placed over this axis according to Scale, considering the total length within the controller space limits. Standard deviations are assigned as half the distance between centre and origin (major axis) and the intersection point with the controller space along its perpendicular (minor axis). Figure ~\ref{fig:user_pref} includes a graphical representation of this mechanism for three different Similarity and Scale configurations on the same compliance profile.

\section{Automated VIC Solution Generation}\label{sec:LfDVIC_AutomatedSol}

\sepfootnotecontent{UmaxMin}{For the implementation of the optimisation problem, the minimisation of $\InputVec_{max}$ makes it non-convex due to constraint~(\ref{eq:LfDVIC_LMI_Operational}a). Therefore, $\InputVec_{max}^2$ is used as the minimisation objective instead, as it implies the minimisation of $\InputVec_{max}$.}

Controller solution assessment as presented in the previous Section only regards about determining whether the tuple of parameters that describes the VIC fulfils a set of conditions or not. As the method seeks one controller solution to be used in the VIC, condition assessment needs to be formulated in order to provide a score that determines the overall suitability of the solution. Hence, in this work, for each controller solution \VICSol, a complete suitability score \VICSolScore is obtained as the sum of a score on safety $\VICSolScore^{\text{Safety}}$ and performance $\VICSolScore^{\text{Perf.}}$. Considering the definition of safety conditions given in this work, its corresponding score needs to characterise the narrower allowable region in the controller solution space, i.e. how much \textit{safer} is \VICSol among all the solutions. Hence, making use of the convexity property of LMIs, safety conditions are stated into a convex optimisation problem to minimise squared limit control effort $\InputVec_{max}^2$ (with a maximum value $\overline{\InputVec_{max}}^2$) as follows~\sepfootnote{UmaxMin}:
\begin{equation} \label{eq:LfDVIC_opt_problem}
	\begin{split}
		&\text{For} \\
		&\text{find} \\
		&\text{minimising} \\
		&\text{subject to} \\
		&\text{given}
	\end{split}
	\quad 
	\begin{split}
		& \StateMatDisc_i\;, \gainVICDisc_i\quad \forall i=1,...,\NumVertex \hfill \\
		& \LyapMatrix \\
		& \InputVec_{max}^2  \\
		& \text{\eqref{eq:LfDVIC_LMI_Stab}, (\ref{eq:LfDVIC_LMI_Operational}a,b), \eqref{eq:LfDVIC_LMI_init_state}, \eqref{eq:LfDVIC_LMI_Dstab_OS}}	\\
		& \StateVec(0), \, \Delta p_{max},  \, \overline{\PercOverShoot},
	\end{split}
\end{equation}
such that $\VICSolScore^{\text{Safety}}$ is obtained by mapping $\InputVec_{max}$ in $[0,\overline{\InputVec_{max}}]$ to $[0,1]$. Performance score $\VICSolScore^{\text{Perf.}}$ is directly assigned to the output from User Preference mechanism. Therefore, the only values to be a-priori defined for controller assessment are a maximum control effort limit $\overline{\InputVec_{max}}$, and $\StateVec(0), \, \Delta p_{max},$  and $\overline{\PercOverShoot}$ for the problem stated in~\eqref{eq:LfDVIC_opt_problem}, and Similarity and Scale for performance condition. 

\begin{figure}[tbp]
	\centerline{\includegraphics[width=\linewidth]{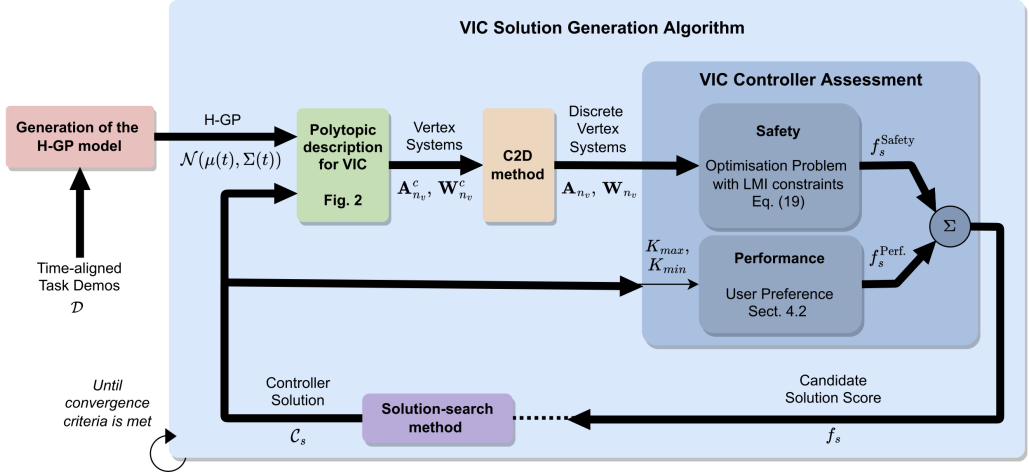}}
	\caption{Complete scheme of the automated generation of controller solutions for LfD-based VIC description, based on proposed condition assessment.}
	\label{fig:LfDVIC_Solution_Generation}
\end{figure}

Hence, the complete generation of the most suitable controller solution $\VICSol^{*}$ to define the VIC can be described. The solution-search method iteratively provides controller solution candidates \VICSol within the limits given to each parameter. Additionally, to obtain the desired behaviour by the LfD technique described in~\eqref{eq:LfDVIC_K_scaling}, constraint $\StiffVIC_{max}>\StiffVIC_{min}$ is imposed at this step. Then, its polytopic description is obtained as described in Sect.~\ref{sec:LfDVIC_PolytopicForm}, considering the H-GP model that embeds task demonstrations. The discrete-time form of vertex systems are used for LMI constraints on the problem regarding safety conditions assessment stated in~\eqref{eq:LfDVIC_opt_problem} to obtain $\VICSolScore^{\text{Safety}}$. Simultaneously, performance is assessed directly on \VICSol providing a performance score $\VICSolScore^{\text{Perf.}}$. Together they compound the suitability score \VICSolScore, which is provided to the solution-search method to generate next candidates towards the most suitable one, until the convergence criteria is met, i.e. $\mathcal{C}_s \approx \mathcal{C}_s^*$. Figure~\ref{fig:LfDVIC_Solution_Generation} graphically depicts the complete process. 

% INTERESTING TO PUT SOMEWHERE - Before the beginning of the solution generation, all the necessary parameters must be provided. From the H-GP data, $x_0$ is obtained through the mean reference trajectory, and $\Delta p_{max}$ is defined as the \textcolor{red}{minimum} difference between the confidence interval (95\%) and the mean reference. 

%\noindent It should be noted that for the implementation of the LMI-problem, the minimisation of $\InputVec_{max}$ makes the optimisation problem non-convex. Hence, $\InputVec_{max}^2$ is used as the minimisation objective instead as it implies the minimisation of $\InputVec_{max}$.

%\begin{algorithm}[!tbp]
%	\input{./Content/Chapter_LfDVIC/Algorithms/tuning_VIC_alg}
%\end{algorithm}

\section{Validation Experiments} \label{sec:LfDVIC_Validation}

\sepfootnotecontent{Kinova}{\textsc{Gen3} robotic manipulators by
	 \Kinova\url{https://www.kinovarobotics.com/product/gen3-robots}}

\begin{figure*}[!t]
	\centerline{\includegraphics[width=\linewidth]{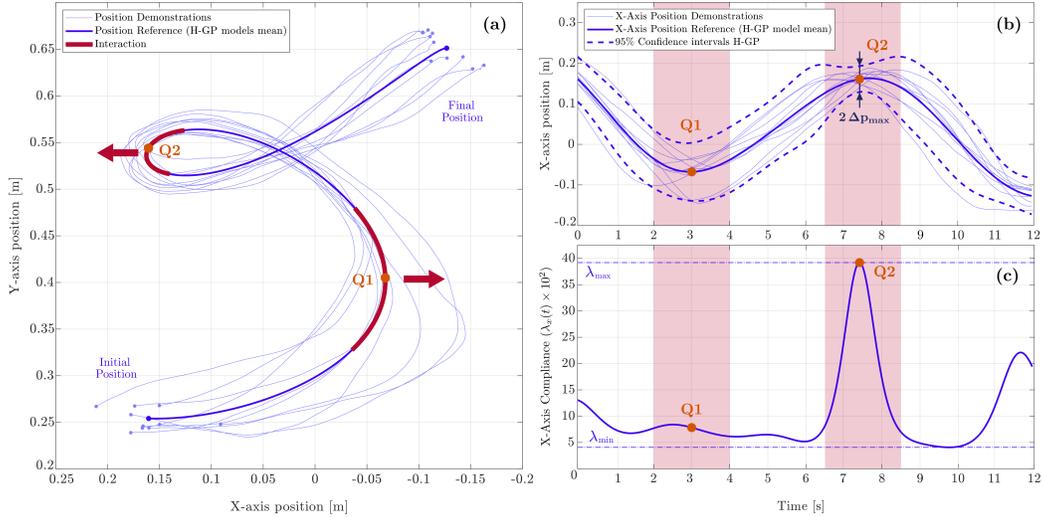}}
	\caption{Demonstrations of the validation trajectory and generated reference trajectory (a) together with the H-GP mean and confidence intervals for the X axis (b) and the corresponding compliance profile (c). Regions where a constant virtual force of 50[N] is applied in opposite directions are highlighted in the reference trajectory in (a) and represented as shadowed areas in (b) and (c), together with the points Q1 and Q2 where compliancy is evaluated.}
	\label{fig:LfDVIC_wipping_gp_compliance}
	%\vspace{-2mm}
\end{figure*}

\begin{figure*}[!tbp]
	\centerline{\includegraphics[width=0.85\linewidth]{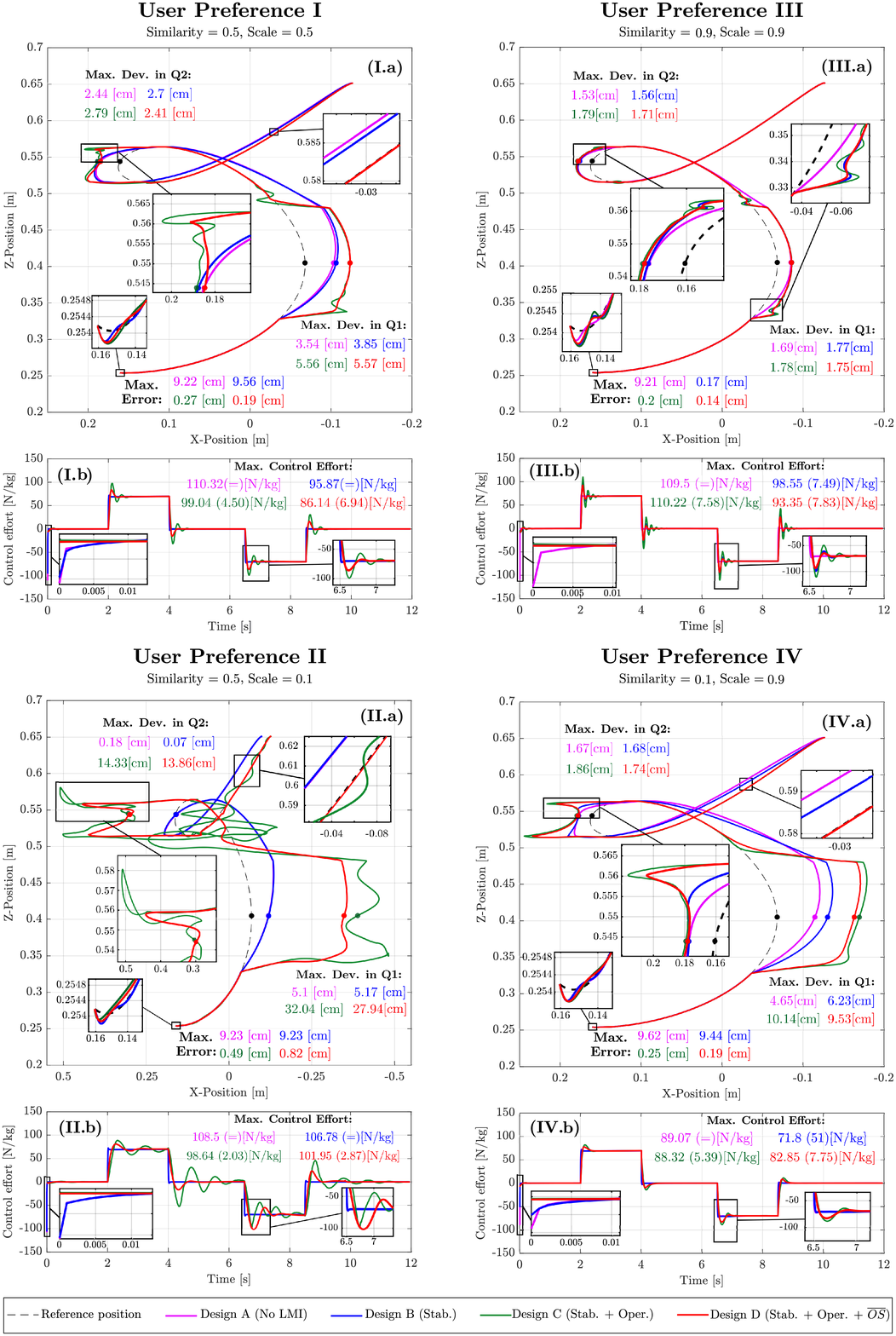}}
	\caption{Validation trajectory execution and control effort evolution using the VIC solution of each User Preference - Design combination. For User Preference N, Figure N.a represents the end-effector position together with the maximum absolute trajectory error at the beginning of the task and deviations at Q1 and Q2, and Figure N.b depicts the evolution of the control effort and its maximum absolute values along the task and before force application (in pnthesis)  annotated.}
	\label{fig:LfDVIC_validation_trajectory}
\end{figure*}

The method presented in this paper is validated on a trajectory tracking task using a \textsc{\Kinova Gen3} robotic manipulator~\sepfootnote{Kinova}. Trajectory consist on a planar ($2$-DoF) ``wiping" movement along a surface, which is embedded in H-GP models (one per axis) using a set of $10$ user-guided demonstrations as shown in Fig.~\ref{fig:LfDVIC_wipping_gp_compliance}a, obtained according to the details provided in App.~\ref{appendix:LfDVIC_HGPGen}. Notice that initial positions for demonstrations are randomly generated (within a given interval) in order to avoid null variability (and consequently highest stiffness value) at the beginning of the task. To better analyse the behaviour of different controller solutions, VIC is applied only for $X$-axis control, while (constant) IC described in Table~\ref{table:LfDVIC_constant_controllers} are used for remaining DoFs. From the H-GP covariance, represented in Fig.~\ref{fig:LfDVIC_validation_trajectory}b by the $95\%$ confidence intervals, the compliance profile determined by $\lambda(t)$ is obtained (Fig.~\ref{fig:LfDVIC_validation_trajectory}c). Based on it, to evaluate compliant behaviour along the task, a constant force is virtually introduced during trajectory execution twice: in a high compliance (low stiffness) region between 2 and 4[s] and in a low compliance (high stiffness) region between 6.2 and 8.2[s]. In each of these regions, the compliant behaviours will be evaluated at certain point, namely Q1 and Q2. Force magnitude is equal in both cases but it is first applied in the negative axis direction and then in the opposite one.

\begin{table}[!t]
	\centering
	\footnotesize
	\begin{tabular}{cccc}
DoF Type        & $H$ [kg] & $K$ [N/m] & $D$ [N$\cdot$s/m]  \\ \hline
Translational   & 2        & 10000     & \multirow{2}{*}{$2\cdot0.8\, \sqrt{K\cdot H}$} \\
Rotational      & 1        & 1500      &
\end{tabular}

	\caption{Constant impedance controller gains for \textsc{KINOVA Gen3} manipulator. Considering it imposes a  2nd-order dynamic behaviour, $D$ is computed as a function of $K$ and $H$ with a damping ratio of 0.8.}
	\label{table:LfDVIC_constant_controllers}
\end{table}

\begin{table*}[!t]
	%\hspace{1mm}
\begin{subtable}[h]{\textwidth}
\centering
\footnotesize
\begin{tabular}{cccccc}
\multirow{2}{*}{Design} &\multirow{2}{*}{$f_s^{\text{User}}$} &  $\InputVec_{max}$ & $K_ {max}$  & $K_{min}$  &  $D$      \\
                      &                                     &   [N/kg]   & [N/m]       &   [N/m]    &  [N$\cdot$s/m]  \\ \hline
A   & 7.81 $\times10^{-5}$  & -      & 4999 &  2086  & 2497 \\ 
B  & 6.68 $\times10^{-3}$  & -      & 4940 &  2067  & 2169  \\ 
C & 3.8  $\times10^{-3}$  & 4.61   & 5019 &  2181  & 71 \\ 
D  & 7.24 $\times10^{-2}$  & 6.94   & 4803 &  1987  & 157 \\ 
\hline 
\end{tabular}
\caption{User Preference I: Similarity = 0.5, Scale = 0.5}
\label{table:LfDVIC_controller_solution_user_I}
\end{subtable}
%\hspace{4mm}
\begin{subtable}[h]{\textwidth}
	\centering
	\footnotesize
	\begin{tabular}{cccccc}
\multirow{2}{*}{Design} &\multirow{2}{*}{$f_s^{\text{User}}$} &  $\InputVec_{max}$ & $K_ {max}$  & $K_{min}$  &  $D$      \\
                      &                                       &   [N/kg]   & [N/m]       &   [N/m]    &  [N$\cdot$s/m]  \\ \hline
A   & 9.02 $\times10^{-4}$ & -     & 990  &  402  &  2455 \\ 
B  & 1.43 $\times10^{-4}$ & -     &  997 &  409  &  2416 \\ 
C & 3.13 $\times10^{-2}$ & 2.23  &  902 &  343  &  17  \\ 
D  & 4.08 $\times10^{-3}$ & 2.87 &  988 &  431  & 65  \\ 
\hline 
\end{tabular}
	\caption{User Preference II: Similarity = 0.5, Scale = 0.1}
	\label{table:LfDVIC_controller_solution_user_II}
\end{subtable}
%\hspace{4mm}
\begin{subtable}[h]{\textwidth}
\centering
\footnotesize
\begin{tabular}{cccccc}
\multirow{2}{*}{Design} &\multirow{2}{*}{$f_s^{\text{User}}$} &  $\InputVec_{max}$ & $K_ {max}$  & $K_{min}$  &  $D$      \\
                      &                                       &   [N/kg]   & [N/m]       &   [N/m]    &  [N$\cdot$s/m]  \\ \hline
A  & 1.12$\times10^{-5}$ &  -& 9018 &  7671 &  2478 \\ 
B  & 3.27$\times10^{-6}$ & - & 8986 &  7687 &  139  \\ 
C  & 2.35$\times10^{-2}$ &  7.58 & 7814 &  7759 &  74  \\ 
D  & 1.21$\times10^{-2}$ &  7.83 & 8181 &  7847 &  177  \\ 
\hline 
\end{tabular}
\caption{User Preference III: Similarity = 0.9, Scale = 0.9}
\label{table:LfDVIC_controller_solution_user_III}
\end{subtable}
%\hspace{1mm}
\begin{subtable}[h]{\textwidth}
\centering
\footnotesize
\begin{tabular}{cccccc}
\multirow{2}{*}{Design} &\multirow{2}{*}{$f_s^{\text{User}}$} &  $\InputVec_{max}$ & $K_ {max}$  & $K_{min}$  &  $D$      \\
                      &                                       &   [N/kg]   & [N/m]       &   [N/m]    &  [N$\cdot$s/m]  \\ \hline
A  & 5.66$\times10^{-5}$  & - & 8961 & 703 & 2016 \\ 
B  & 1.92$\times10^{-3}$  & - & 9261 & 721 & 1154  \\ 
C  & 5.68$\times10^{-2}$  & 5.55 & 7467 & 599 & 122  \\ 
D  & 2.47$\times10^{-2}$  & 7.75  & 8002 & 621 & 175\\ 
\hline 
\end{tabular}
\caption{User Preference IV: Similarity = 0.1, Scale = 0.9}
\label{table:LfDVIC_controller_solution_user_IV}
\end{subtable}
\caption{Controller solutions for each User Preference - LMI Design combination, together with their corresponding value of User Preference score ($f_{S}^{\text{User}}$) and limit control effort ($\InputVec_{max}$).}
\label{table:LfDVIC_general_controller_wipping}
\end{table*} 

\begin{figure*}[!t]
	\centerline{\includegraphics[width=\linewidth]{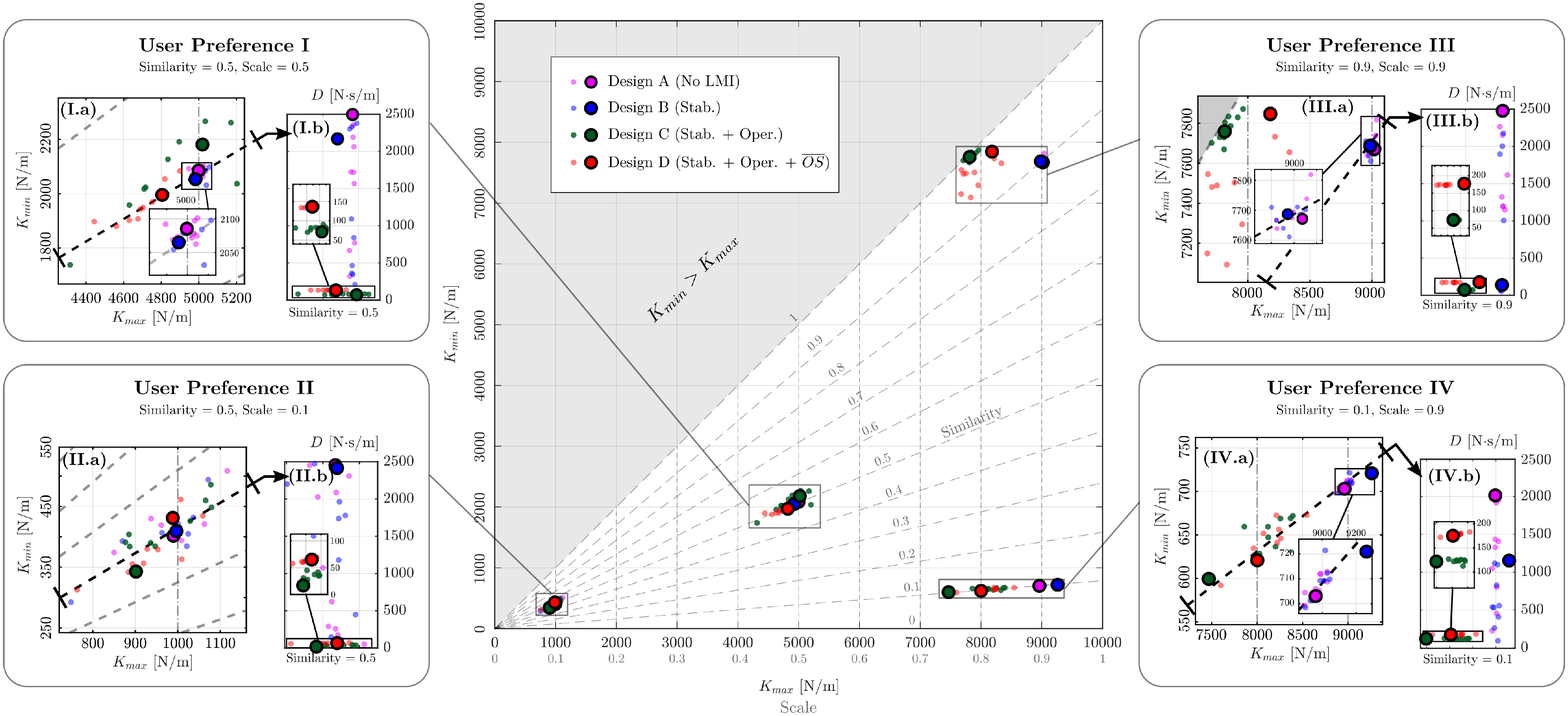}}
	\caption{Controller solution space with the ones generated for each User Preference - Design combination. Central figure represents stiffness plane ($K_{max}$-$K_{min}$) and, for User Preference N, Figure N.a represents a zoom over the region with all the obtained solutions. Figure N.b represents the (vertical) projection over the corresponding Similarity plane to visualize $D$ values. Chosen controller solutions are represented by larger size dots.}
	\label{fig:LfDVIC_wipping_controller_space}
\end{figure*}

 Prior to the generation of VIC solutions, the different parts that compound the method must be defined according to the task and platform. First, given parameters to the optimisation problem~\eqref{eq:LfDVIC_opt_problem} have to be determined. Initial state $\StateVec(0)$ is obtained from the reference trajectory generated by H-GP models, considering that the robot starts at the initial point of the reference trajectory with null velocity. Remaining parameters are assigned by the user, in this case $\Delta p_{max}$ is considered to be the minimum difference between the $95\%$ confidence bounds and reference trajectory from the H-GP model(Fig.~\ref{fig:LfDVIC_wipping_gp_compliance}b), having a value of 3.19[cm]. Considering experience on the real platform, $\overline{\InputVec_{max}}=10$[N/kg] and $\overline{PO} = 5\%$. For the solution-search method, bounds on \VICSol parameters are given to avoid noise amplification effects on the real platform such that $K_{min}$ and $K_{max}$ have been limited to $[0,10000]$~[N/m] and $D$ to $[0,2500]$~[N$\cdot$s/m]. The VIC description is completed with $H = 2$~[kg], determined according to robot hardware~\cite{Dietrich2021}. The discretisation process of vertex systems' matrices is performed considering signal processing on the platform, hence using Zero-order hold with a sampling time $T_s=1$ [ms] (as control-loop frequency is $1$~[kHz]).

\sepfootnotecontent{ProjectWeb}{Project webpage:
	\url{http://www.iri.upc.edu/groups/perception/\#LMI_VIC_LfD}. \label{footnote:ProjectWeb}}

\sepfootnotecontent{DesignOpti}{As Designs A and B do not incorporate LMI constraints with $\InputVec^2$, the norm of solution matrix \LyapMatrix is internally used as minimisation objective, but is not assigned to $\VICSolScore^{\text{Safety}}$.}

Controller solutions are obtained for four sets of User Preferences describing different behaviours, labeled as I-IV. For each one, four designs have been performed to analyse the effect of introducing each LMI constraint into the optimisation problem~\eqref{eq:LfDVIC_opt_problem}. Thus, Design A has no constraints (only considers User Preference), Design B incorporates stability constraint~\eqref{eq:LfDVIC_LMI_Stab}, Design C adds the limitations on position error and control effort~\eqref{eq:LfDVIC_LMI_Operational}, and finally Design D incorporates the maximum overshooting constraint, i.e. the complete optimisation problem as stated in~\eqref{eq:LfDVIC_opt_problem}~\sepfootnote{DesignOpti}. Hence, for each User Preference - Design combination, $10$ solution controllers have been obtained to evaluate convergence under different initialisation. The chosen ones for task execution  those with $\VICSolScore^{Perf.}$ closest to its mean. Under the implementation detailed in App.~\ref{appendix:LfDVIC_method_impl} solutions depicted in Figure~\ref{fig:LfDVIC_wipping_controller_space} at the controller parameter space are obtained, and Table~\ref{table:LfDVIC_general_controller_wipping} summarises the details of chosen ones for execution. These results can be replicated using the demo version of the complete method available in the project webpage~\sepfootnote{ProjectWeb}.

%Additional details on algorithm implementation and execution can be found in the extended appendix available in the dedicated project webpage~\footnote{\label{lab:webpage} \url{http://www.iri.upc.edu/groups/perception/\#LMI_VIC_LfD}}, where a demo version of the complete algorithm for the validation task is also available.

\subsection{Results} \label{sec:LfDVIC_validation_results}

Due to the chosen validation experiment, some controllers are obtained with minimal or even no constraints, which makes not possible to ensure a safe execution of the task beforehand regarding both the robot, e.g. reaching joint position and force limits, and the environment, e.g. colliding with the surroundings in case an unstable behaviour arises. Therefore, trajectory execution has been carried out in the physic-based simulator \Simscape within the \Matlab programming ecosystem. As in the real platform, implemented control strategy renders~\eqref{eq:LfDVIC_imp_relation} through the well-known inverse dynamic approach to compensate for non-linear effects, also replicating its discrete-time execution. All trajectory executions are summarised into a video provided as supplementary material and also available in the project webpage$^\text{\ref{footnote:ProjectWeb}}$.

%Additional details on algorithm implementation and execution  given in the extended appendix available in the project webpage~\footnote{\label{lab:webpage} \url{http://www.iri.upc.edu/groups/perception/\#LMI_VIC_LfD}}.
%All the obtained controllers as well as a demo version of the complete algorithm for the validation task is available in the dedicated project git page~\footnote{Git webpage: \url{https://gitlab.iri.upc.edu/asanmiguel/LMI_design_VIC_toolbox}}.

Figure~\ref{fig:LfDVIC_validation_trajectory} shows trajectory executions of the controller solutions obtained for each User Preference - Design combination under the introduction of virtual forces. User Preference I sets a desired compliant behaviour centered in both Scale and Similarity ranges. Both controllers generated from Designs A and B present a highly damped behaviour along the task with respect to Designs C and D, due to the high values of $D$ gain. Those designs do not include LMI constraints able to fix its value, and, for that reason, the solution generation method randomly assigns it within the allowed range, which can be seen in Fig.~\ref{fig:LfDVIC_wipping_controller_space}.I.b. As Fig.~\ref{fig:LfDVIC_validation_trajectory}.I.a shows, this produces a ``slow'' reaction when force is applied, which is most noticeable in the first region, where the robot is driven gradually away from the trajectory. This also affects the behaviour when force disappears, maintaining a high position tracking error for longer than solutions from Designs C and D. Regarding control effort (Fig.~\ref{fig:LfDVIC_validation_trajectory}.I.b) this leads to an unbounded value that takes its highest value (considering the no-force sections of the trajectory) at the initial point. It should be mentioned that having a non-zero control effort at the beginning of the task is a consequence of the procedure to obtain the reference velocity from the position mean of the H-GP model, which generates an initial non-zero velocity reference. Therefore, as the robot starts stalled, a control effort is generated to reach reference velocity. This does not happen with Designs C and D as they introduce a condition that provides an upper bound of the control effort (again, under no external force) considering $\StateVec(0)$. Between these two designs, the most recognisable difference is the tracking behaviour (Fig.~\ref{fig:LfDVIC_validation_trajectory}.I.a) when the force is applied and fades out: while the controller from Design C presents a noticeable oscillatory response, the one from Design D presents a softened reaction, which becomes more noticeable at the beginning of second force application. This is an effect of minimising the maximum control effort, leading to low values of $D$ that still fulfil the maximum deviation condition. By introducing the maximum overshooting condition in Design D, although the maximum control effort is minimised, $D$ is set to a higher value in order to reduce the oscillatory behaviour. This can also be observed for the control effort evolution in Fig.\ref{fig:LfDVIC_wipping_controller_space}.I.b. Note that in both Designs C and D control effort evolution present lower values at the beginning (fulfilling the bounds defined by design in Table~\ref{table:LfDVIC_controller_solution_user_I}), obtaining the lowest one with Design D solution. Also the max. error (evaluated before first force applications) remains below the value of $\Delta p_{max}$ chosen from the H-GP model in both designs.

Similar conclusions can be drawn for the remaining User Preferences with some particularities on each case. From User Preference I to II Scale is reduced to $0.1$ while Similarity is maintained at $0.5$. This generates solutions closer to the lower limits of the stiffness domain with Similarity again centered in the interval. Thus, high and low compliance behaviours are quite similar but still maintain a substantial contrast, represented by the maximum deviations in Q1 and Q2. Regarding the differences between designs, controllers from A, B and D present similar characteristics than the ones obtained for User Preference I. For C, a lower $D$ together with the low values of $K_{min}$ and $K_{max}$ due to Scale causes an erratic behaviour on the second force application, as it can be observed in Fig.~\ref{fig:LfDVIC_validation_trajectory}.II.b. For User Preference III, both Similarity and Scale are increased to $0.9$, which translates into high and similar values for $K_{max}$ and $K_{min}$. Thus, the difference between maximum deviations under different compliance levels for this case is the smallest (between $0.01$ and $0.14$~[cm]). In this case, all the controllers behave similarly for the tracking task (Fig.~\ref{fig:LfDVIC_validation_trajectory}.III.a) but there exist still differences during force application, being the reaction from Design D controller faster than A but with less oscillations than D. Notice that in this case, controller obtained with Design B has almost the same behaviour as the one from Design D, due to their similar $D$ gain. Again, as with Design A, this gain is not fixed by any constraint and its value is randomly assigned (Fig.~\ref{fig:LfDVIC_validation_trajectory}.III.b). Finally, from User Preference III to IV only Similarity is changed to $0.1$, which generates a variant controller solution that presents the greatest deviations differences between Q2 and Q1 from $3$ to almost $9$~[cm]. As in User Preference I, Designs A and B do present a damped tracking behaviour (Fig.~\ref{fig:LfDVIC_validation_trajectory}.IV.a) that slows its response under force application, leading to high trajectory tracking errors. In this case, controllers from both Designs C and D have a quite similar response although an oscillatory behaviour is still present in the former one, being most noticeable at the beginning of the first force application. Finally note that both max. control effort (Tables~\ref{table:LfDVIC_controller_solution_user_I}~-~\ref{table:LfDVIC_controller_solution_user_IV}) and max. deviation bounds are fulfilled by Designs C and D for all User Preferences.

\section{Case Study} \label{sec:LfDVIC_case_study}

\begin{figure}[!t]
	\centerline{\includegraphics[width=\linewidth]{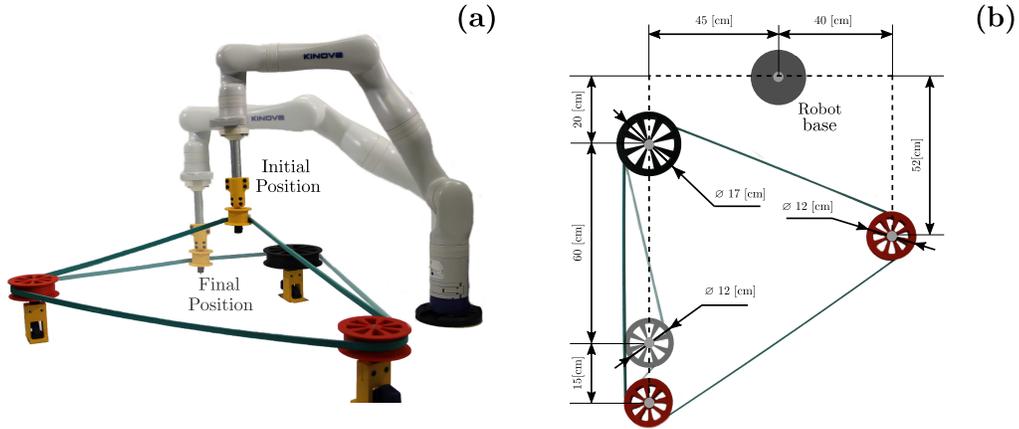}}
	\caption{Belt Drive unit setup for Case Study. Figure a depicts the initial and final (shadowed) positions of the KINOVA manipulator and belt configurations. Figure b includes the distribution of the pulleys w.r.t. the robot base, including the idler pulley (shadowed) for the first one-off modification scenario.}
	\label{fig:LfDVIC_pulley_setup}
\end{figure}

Finally, the method proposed in this work is evaluated on a real case study. Following the Assembly Challenge on the Industrial Robotics Category in World Robot Summit 2018~\cite{yokokohji2019assembly}, the task consists on looping a belt over the last pulley in a belt drive unit (Fig.~\ref{fig:LfDVIC_pulley_setup}). Within the Challenge, this task aimed at evaluating the manipulation of flexible objects on a complex trajectory while interacting with other elements. Hence, the method presented in this work is used to provide a task definition through demonstrations such that a VIC is generated to reduce stress over the belt during execution w.r.t. a constant rigid controller. Moreover, as in the Challenge, slight one-off modifications on the setup are used to evaluate the adaptability to unseen new scenarios, which is tackled through different User Preference configurations. In the first scenario an idler pulley is introduced (depicted also in Figure~\ref{fig:LfDVIC_pulley_setup}b), and in the second one the belt used in the nominal scenario is substituted by a stiffer one.

%In the Challenge, this task was the last step of a belt drive unit assembly, a common mechanism in many manufactured machinery products. This part aimed at evaluating the manipulation of flexible objects on a complex trajectory while interacting with other elements (pulleys). In this case, we want to apply the method presented in this paper in order to (i) make use of human knowledge on the task and (ii) reduce stress over the belt w.r.t. a constant rigid controller. Moreover, as in the challenge, we present slight one-off modifications to evaluate the adaptability of the method to unseen new scenarios using the user preference mechanism. The first scenario consists on the addition of an idler pulley (depicted also in Figure~\ref{fig:pulley_setup}b), and the second one on changing the belt on the nominal setup to a stiffer one. 

\begin{figure*}[!t]
	\centerline{\includegraphics[width=\linewidth]{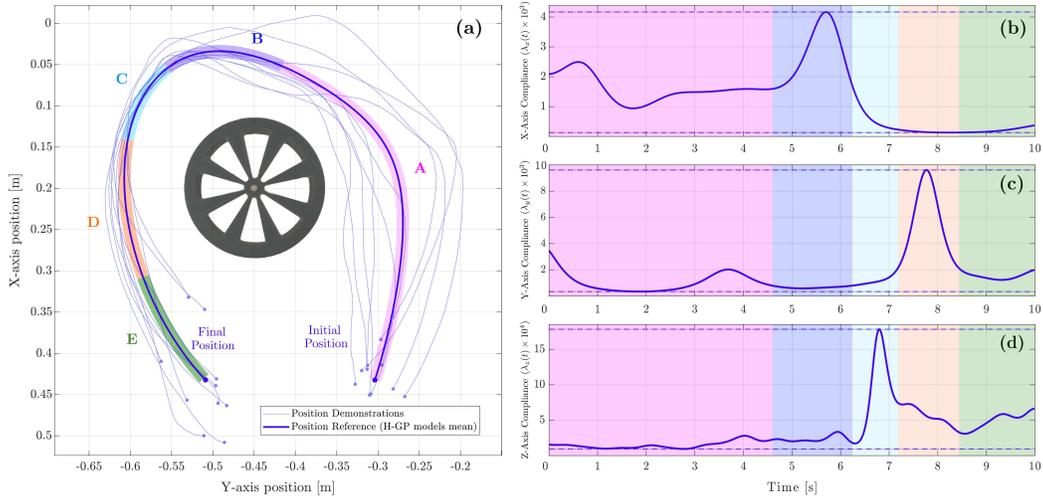}}
	\caption{Demonstrations of the looping task trajectory and generated reference trajectory (a) together with the corresponding compliance profiles in $X$ (b), $Y$ (c) and $Z$ (d)-axes. The different task regions are highlighted in the reference trajectory in (a) and represented as shadowed areas in (b-d). Region A corresponds to the approach movement from the initial position to the pulley (low/medium stiffness in all three axis), Regions B to D to the stiffness maximum in $X$ (beginning of the looping action), $Z$ (to ensure downwards movement is performed to engage belt) and $Y$-axes (to avoid colliding the pulley), respectively. Region D is for the last movement away from the pulley to the final position.}
	\label{fig:LfDVIC_pulley_gp_compliance}
\end{figure*}

\begin{table}[!t]
	\begin{subtable}[h]{\textwidth}
\centering
\footnotesize
\begin{tabular}{cccccc}
\multirow{2}{*}{Axis}    & \multirow{2}{*}{$f_s^{\text{User}}$} & $\InputVec_{max}$ & $K_ {max}$  & $K_{min}$  &  $D$      \\
                         &                                      & [N/kg]    & [N/m]       &   [N/m]    &  [N$\cdot$s/m]  \\ \hline
X   & 6.14$\times10^{-4}$  & 9.38    & 4912              & 1181              & 128                         \\
Y   & 1.52$\times10^{-3}$  & 4.69    & 4861              & 1169              & 160                         \\
Z   & 1.24$\times10^{-3}$  & 4.86    & 5060              & 1241              & 158                  
\end{tabular}
\caption{Nominal Scenario Controllers: Similarity = 0.3, Scale = 0.5}
\label{tab:LfDVIC_controller_solution_pulley_nominal}
\end{subtable}
\begin{subtable}[h]{\textwidth}
\centering
\footnotesize
\begin{tabular}{cccccc}
\multirow{2}{*}{Axis} & \multirow{2}{*}{$f_s^{\text{User}}$} & $\InputVec_{max}$  & $K_ {max}$  & $K_{min}$  &  $D$      \\
                      &                                      &  [N/kg]    & [N/m]       &   [N/m]    &  [N$\cdot$s/m]  \\ \hline
X   & 2.22$\times10^{-4}$  & 9.39                   & 4934              & 4262              & 138                   \\
Y   & 1.08$\times10^{-3}$  & 4.16                   & 4989              & 4130              & 184                   \\
Z   & 3.78$\times10^{-2}$  & 0.4                    & 4646              & 4621              & 428                  
\end{tabular}
\caption{Idler Pulley Scenario Controllers: Similarity = 0.9, Scale = 0.5}
\label{tab:LfDVIC_controller_solution_pulley_idler}
\end{subtable}
\begin{subtable}[h]{\textwidth}
\centering
\footnotesize
\begin{tabular}{cccccc}
\multirow{2}{*}{Axis}   & \multirow{2}{*}{$f_s^{\text{User}}$} & $\InputVec_{max}$  & $K_ {max}$  & $K_{min}$  &  $D$      \\
                        &                                      & [N/kg]     & [N/m]       & [N/m]      &  [N$\cdot$s/m]  \\ \hline
X   & 2.75$\times10^{-3}$ & 12.46      & 8674      & 2067              & 183                   \\
Y   & 2.75$\times10^{-2}$ & 8.15       & 8081      & 2089              & 208                   \\
Z   & 2.01$\times10^{-2}$ & 8.5        & 8148      & 1874              & 205                  
\end{tabular}
\caption{Stiffer Belt Scenario Controllers: Similarity = 0.3, Scale = 0.9}
\label{tab:LfDVIC_controller_solution_pulley_stiffer}
\end{subtable}
\caption{Controller solutions for every axis in each case study scenario, together with their corresponding value of user preference term ($f_{S}^{\text{User}}$) and limit control effort ($\InputVec_{max}$).}
\label{tab:LfDVIC_general_controller_pulley}

\end{table} 

In these experiments, independent VICs are used for the translation control of the robot end-effector, namely in $X$, $Y$ and $Z$-axis, while orientation ICs remain as in validation experiments (Table~\ref{table:LfDVIC_constant_controllers}). The 
 H-GP models are obtained using $10$ human guided demonstrations looping the belt on the nominal setup, starting from random initial conditions within an interval. Figure~\ref{fig:LfDVIC_pulley_gp_compliance}a shows obtained reference trajectory together with used demonstrations, and Figs.~\ref{fig:LfDVIC_pulley_gp_compliance}b-d show the compliance ($\lambda(t)$) profile for $X$,$Y$ and $Z$-axis, respectively. In all Figs.~\ref{fig:LfDVIC_pulley_gp_compliance}a-d task is divided into a set of regions (A to E) based on compliance profiles evolution, as detailed on their caption.

\begin{figure*}[!t]
	\centerline{\includegraphics[width=\linewidth]{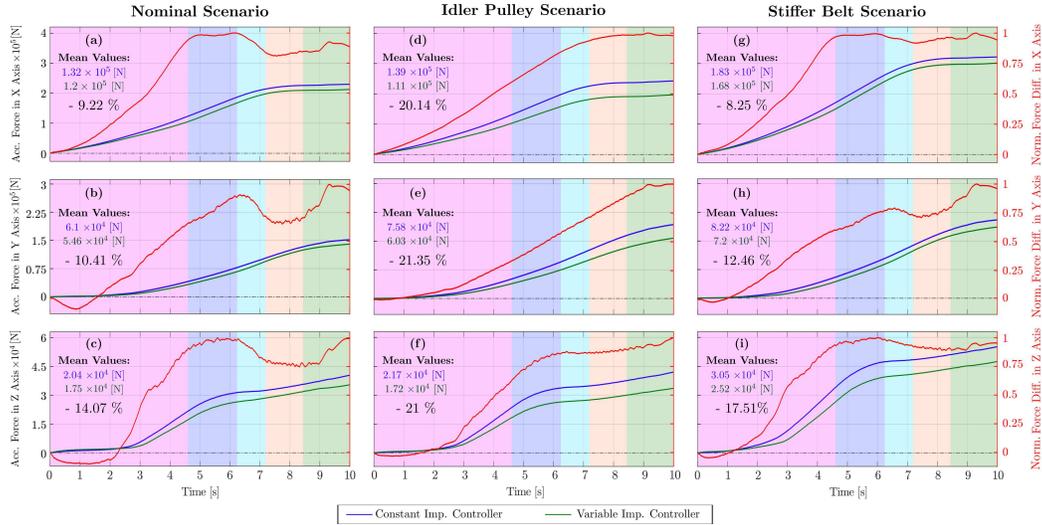}}
	\caption{Accumulated Force of the Constant and Variable Impedance Controller solutions together with the Normalized Difference (w.r.t. the maximum value) between them for $X$, $Y$ and $Z$ axis on the three scenarios for the Belt Unit Case Study: Nominal (a-c), Idler Pulley (d-f) and Stiffer Belt (g-i). For each one, the mean accumulated force values and the reduction/increase that the value for the variable controller represents over the constant one have been also included.}
	\label{fig:LfDVIC_pulley_results}
\end{figure*}

The automated VIC generation definition is given from the same considerations from validation experiments. Hence, from generated H-GP models, the values of $\Delta p_{max}$ are 3.12~[cm] for $X$-axis, 2.03~[cm] for $Y$-axis and 0.55~[cm] for $Z$-axis. Note that the same Scale and Similarity values are used for all axis in order to ease interpretation of the results. Thus, controller solutions on Table~\ref{tab:LfDVIC_general_controller_pulley} are obtained, and the results of their execution in the real platform are depicted in Fig.~\ref{fig:LfDVIC_pulley_results}. Recalling the aim of using VIC to reduce belt stress during the looping task, the accumulated force exerted on the end-effector along all the directions is represented. A video with all the executions using generated VICs is available as supplementary material and can be also found in the dedicated webpage$^\text{\ref{footnote:ProjectWeb}}$.

\subsection{Nominal scenario}  \label{sec:case_study_nominal} 

\sepfootnotecontent{SimultControl}{Simultaneously controlling orientation and position might concur in some deviation from the desired behaviour from one in the other, as reported in~\cite{Dietrich2021}. In this case, the constant impedance controller for orientation could be implemented by imposing its behaviour over the variable one when its stiffness is lower, which can be accentuated due to the initial acceleration caused by the non-zero velocity. ~\label{footnote:simult_orient_control}}

In this scenario, no changes are made from the setting used in task demonstrations. Thus, the objective is to find a variable impedance controller solution that reduces the overall force required to perform the task. For this initial setup, we have chosen a Scale value of 0.5 to obtain a maximum stiffness value in the middle of allowed range, and a Similarity value of 0.3 to emphasize the effect of variant behaviour, i.e. increase the difference between $K_{max}$ and $K_{min}$. As it can be seen in Fig.~\ref{fig:LfDVIC_pulley_results}a-c, in all three axis, mean accumulated force is reduced w.r.t. the constant controller strategy between a 9\%~($X$-axis) and a 14\%~($Z$-axis). Looking at the normalized force difference, it can be seen how stiffness modulation alters the exchanged force along the task. In the first Region, the difference increases in all axis as the stiffness remains low, even though within the first two seconds the variant controller exerts higher forces in $Y$-axis~(Fig.~\ref{fig:LfDVIC_pulley_results}b) and $Z$-axis~(Fig.~\ref{fig:LfDVIC_pulley_results}c) than w.r.t. the variant approach (min. of -0.1), which might be an effect of the simultaneous control of position and orientation~\sepfootnote{SimultControl}. Then, in Region B, stiffness increase in $X$-axis does not present a noticeable effect in the force difference, but the same phenomena in $Z$-axis produces a reduction of the difference in Region C for all the axis, as the controller becomes stiffer for the end-effector to reach the desired height. At this point, it is worth remarking that the stiffness regulation in each axis propagates to all the other ones as they all affect the belt elongation. In Region D, stiffness peak in $Y$-axis maintains force difference in all axis, i.e. the controller holds stiff to maintain the position in $Y$ at the cost of increasing exchanged force. Finally, accumulated force difference rises as a result of the stiffness reduction in all axes at Region E.

\subsection{Idler pulley scenario} \label{sec:LfDVIC_case_study_idler}

\begin{figure}[!tbp]
	\centerline{\includegraphics[width=0.7\linewidth]{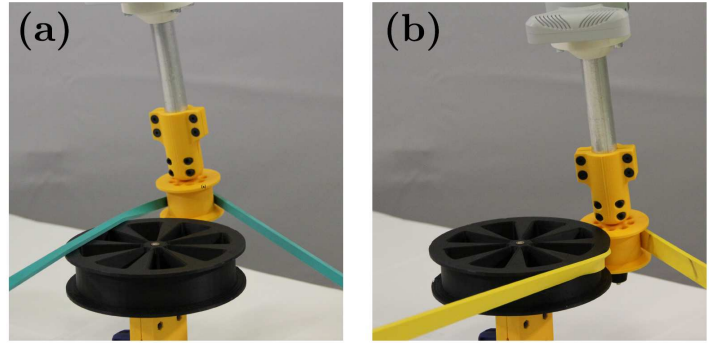}}
	\caption{Task execution errors on the Belt Unit using the nominal controller solutions (Table~\ref{tab:LfDVIC_controller_solution_pulley_nominal}) on the Idler Pulley (a) and Stiffer Belt (b) scenarios.}
	\label{fig:LfDVIC_pulley_errors}
\end{figure}

The first one-off modification consists on introducing a complementary pulley that provides tension and guides the belt, namely idler. It is positioned next to one of the pulleys where belt is looped before starting the task execution (Fig.~\ref{fig:LfDVIC_pulley_setup}b), and so the belt also makes contact with it from the beginning. The first approach is to use the same controller as in the nominal scenario, but, although the task can be accomplished, the belt slides over the upper face of the pulley when lowering the end-effector to the desired height (Fig.~\ref{fig:LfDVIC_pulley_errors}a). This takes place between Regions C and D (7-7.5[s]), just before reaching the maximum stiffness in $Y$-axis. This means that the VIC does not convey enough force to reach a position in $Y$-axis that avoids belt sliding over the pulley. Therefore, from the nominal User Preference, Similarity is increased to 0.9 such that the minimum stiffness gets closer to the maximum one, i.e. a high rigidity is maintained from the beginning of the task. Applying new VICs, task is performed without any issue and the mean accumulated force in all three axis is reduced to values around 20\%. Note that the normalized force difference profile has a more uniform behaviour along the task in comparison with the nominal scenario. This means that the behaviour is closer to an equivalent (in the same range of values) constant controller, due to the high Similarity between $K_{min}$ and $K_{max}$.

\subsection{Stiffer belt scenario} \label{sec:LfDVIC_case_study_stiffer}

For the second scenario, a stiffer belt w.r.t. the nominal one is used. As it can be seen in Fig.~\ref{fig:LfDVIC_pulley_errors}b, VICs generated for the nominal scenario are not able to fulfil the task. Due to the additional rigidity of the stiffer belt, the controller does not provide enough effort to reach the required position before lowering to the desired height and the looping movement is not performed. This happens just after the stiffness peak in $X$-axis (6-6.5[s]), which means that even the maximum stiffness is not enough to counteract the forces from the stiffer belt. Hence, from the nominal User Preference, Scale is increase to 0.9 such that greater values of both $K_{max}$ and $K_{min}$  obtained. This new set of VICs is able to complete the task while reducing the mean accumulated force to values from an 8\% ~($X$-axis) to a 17.5\%~($Z$-axis). The normalized force difference profiles are akin to the nominal scenario ones as both have the same Similarity value. This can be explained through the ratio between maximum and minimum stiffness, which can be seen on Tables~\ref{tab:LfDVIC_controller_solution_pulley_nominal} and \ref{tab:LfDVIC_controller_solution_pulley_stiffer} to be approx. 4 in all axes. Note also that, w.r.t. nominal scenario, the negative difference in the first two seconds (Figs.~\ref{fig:LfDVIC_pulley_results}h,i) is reduced to a min. of -0.05. This might be due to the higher values at the lower stiffness region, which make the VICs less prone to cross-effects from the simultaneous (constant) IC used for orientation. 

%----------
\section{Conclusions and Future Work}\label{sec:LfDVIC_Conclusions}

This paper presents an approach to set conditions on VICs defined through LfD techniques by the off-line tuning of the parameters that complete its description. Hence, using the approach proposed in~\cite{Calinon2010}, a time-variant compliance profile from a set of human-guided task demonstrations is used to determine the stiffness term modulation of the VIC. A solution-search method provides sets of parameter values, namely controller solutions, that are assessed in terms of safety and performance conditions. In this work, first ones are stability, limits on position error and control effort, and maximum overshoot (percentage) in the transient response, which are formulated as LMIs on the polytopic description of the VIC. Regarding performance, the User Preference mechanism allows to assess controller solution w.r.t. user intuition over the task, formalised through an heuristic described by Similarity and Scale. This assessment is formulated to provide a compound suitability index for each controller solution, to be used by the solution-search method in the next iterations. Iteratively repeating this process leads to the most suitable controller solution to define the VIC used in task execution. 

Validation results in simulation show that generated solutions define VICs that (i) fulfil safety conditions, i.e. they are stable, do not surpass design limits on position error and reduce the oscillatory behaviour in the transient response w.r.t. solutions from relaxed assessments, and (ii) reflect a compliant behaviour in line with the inputs to User Preference mechanism that conforms performance assessment. In the case study scenario where a real manipulator is used to loop a belt over a pulley (i) using generated VICs reduces the accumulated total force required to perform the task in all the settings as (ii) User Preference mechanism allows to generate different solutions for the same control architecture fulfilling the same conditions without new task demonstrations.

Future works will explore the application of this method for other LfD approaches, e.g. those that exploit exerted force in human-guided demonstrations to perform position-constrained tasks as in~\cite{Abu-Dakka2018}. Also, controller assessment will be further extended, mainly in the form of LMI constraints, e.g. introducing passivity for human-interaction tasks~\cite{stramigioli2015energy}. 

%This analysis could be used as a first step in a real case scenario to refine the selection of scale and similarity values to greater reduce the accumulated force. For example, as the stiffness increase in $X$-axis (Region B) does not deliver a reduction of the force difference, i.e. increasing rigidity at that point does not involve an increase on the required force to drive the robot end-effector to that point, the scale can be set to a lower value, meaning a more compliant controller over all the trajectory could be used to undertake the task.  

\section*{Acknowledgements}
This work is partially supported by MCIN/ AEI /10.13039/501100011033 and by the "European Union NextGenerationEU/PRTR" under the project ROB-IN
(PLEC2021-007859); and by MCIN/ AEI /10.13039/ 501100011033, Spain, under the project CHLOE-GRAPH (PID2020- 119244GB-I00). Authors also want to thank Adrià Colomè and Edoardo Caldarelli for their comments and help throughout this work.
%\newpage
\section*{Appendix}
\vspace{-10mm}
%\appendix

\begin{appendices}

\section{Proofs of LMI constraints} \label{appendix:LfDVIC_LMI_proofs}

\subsection{Proof of Proposition \ref{prop:LfDVIC_Stab}} \label{appendix:LfDVIC_LMI_Proof_Stab}

According to Lyapunov theory, the equilibrium point $\StateVec(k) = 0$ is stable if there exists a discrete-time candidate function $\LyapCand(\StateMatDisc(k))$ such that $\forall k\geq0$: (i) $V(0) =0\,$; (ii) $V(\StateVec(k))  \geq 0\,,\; \forall\StateVec(k) \neq 0$; and  (iii) $\LyapCand(\StateVec (k+1))-\LyapCand(\StateVec(k)) \leq 0 \, , \forall \StateVec(k)\neq 0\,$. Applying the LMI approach for a generic quadratic candidate function
\begin{equation*}
 \LyapCand(\StateVec(k))=\StateVec(k)^T\cdot\LyapMatrix\cdot\StateVec(k),
\end{equation*} 
from condition (iii), conditions (i) and (ii)  fulfilled iff there exist a solution matrix $\LyapMatrix=\LyapMatrix^T>0$ such that the LMI holds:
\begin{equation}\label{eq:stab_common}
	{\StateMatDisc(k)}\cdot\LyapMatrix\cdot{\StateMatDisc(k)}^T - \LyapMatrix \leq \ZeroMatrix
\end{equation}
Considering the polytopic description of the VIC in Eq.~\eqref{eq:LfDVIC_LPV_state_space} defined by the system evaluated at the convex hull \ConvexHull, Lyapunov stability implies that solution $\LyapMatrix$ is common to all the vertex state matrices $\StateMatDisc_i$, turning~\eqref{eq:stab_common} into~\eqref{eq:LfDVIC_LMI_Stab}. \hfill $\blacksquare$

%-----------------------------------------------------------------------------------------------------------------------------------

\vspace{2mm}
\subsection{Proof of Proposition \ref{prop:LfDVIC_limits}}\label{appendix:LfDVIC_LMI_Proof_limits}

Following \cite{koth1996robust}, considering solution matrix $\LyapMatrix = \LyapMatrix^T > 0$ that shapes an invariant ellipsoid $\Ellipsoid \eqcolon \{ \VarEllipsoid \, | \, \VarEllipsoid^T\cdot \LyapMatrix \cdot \VarEllipsoid < \IdentMat\}$ and its inverse $\InvLyapMatrix$, and the intermediate variable $\GainAnalysisMatrix=\gainVICDisc(k)\cdot\InvLyapMatrix$, condition~(\ref{eq:LfDVIC_limits}a) can be stated through the maximum norm of the control effort:
\begin{subequations}
	\begin{align*}
		\parallel \InputVec(k) \parallel_2^2 &\leq \underset{k\geq 0}{\text{max}} \parallel \InputVec(k) \parallel_2^2 = \underset{k\geq 0}{\text{max}} \parallel \GainAnalysisMatrix \cdot \InvLyapMatrix^{-1} \cdot  \StateVec(k)\parallel_2^2 \\ 
		&\leq \underset{\VarEllipsoid\,\in\,\Ellipsoid}{\text{max}} \parallel \GainAnalysisMatrix \cdot \InvLyapMatrix^{-1} \cdot \StateVec\parallel_2^2 = \overline{\eigenval}(\InvLyapMatrix^{-1/2} \cdot \GainAnalysisMatrix^T\cdot \GainAnalysisMatrix \cdot \InvLyapMatrix^{-1/2}) \leq \InputVec_{max}^2
	\end{align*}
\end{subequations}
Applying Schur lemma and pre/post-multiplying by $[\IdentMat \, \InvLyapMatrix]^T$, leads to~(\ref{eq:LfDVIC_LMI_Operational}a) considering the polytopic description. Similarly, for operational constraint~(\ref{eq:LfDVIC_limits}b), introducing output equation
\begin{equation}
	\PosError(k) = \SelectMatrix \cdot \StateVec(k)
\end{equation}
leads to
\begin{subequations}
	\begin{align*}
		\parallel \PosError(k) \parallel_2^2 &\leq \underset{k\geq 1}{\text{max}} \parallel e_k \parallel_2^2 = \underset{k\geq 1}{\text{max}} \parallel \SelectMatrix \cdot \StateMatDisc \cdot \StateVec(k-1)\parallel_2^2 \\ &\leq \underset{\VarEllipsoid\,\in\,\Ellipsoid}{\text{max}} \parallel \SelectMatrix \cdot \StateMatDisc \cdot \VarEllipsoid\parallel_2^2 = \overline{\eigenval}(\InvLyapMatrix^{1/2} \cdot {(\SelectMatrix \cdot \StateMatDisc)}^T(\SelectMatrix \cdot \StateMatDisc) \cdot \InvLyapMatrix^{1/2}) \leq \Delta p_{max}^2
	\end{align*}
\end{subequations}
Applying the Schur lemma and  pre/post-multiplying by $[ \InvLyapMatrix \, \IdentMat]^T$, leads to~(\ref{eq:LfDVIC_LMI_Operational}a) considering the polytopic description. Condition~\eqref{eq:LfDVIC_LMI_init_state} ensures that the initial state $\StateVec(0)$ belongs to \Ellipsoid by applying the Schur lemma on its definition and pre/post-multiplying its result by $[\IdentMat \, \InvLyapMatrix]^T$. \hfill $\blacksquare$

%-----------------------------------------------------------------------------------------------------------------------------------

\vspace{2mm}
\subsection{Proof of Proposition \ref{prop:LfDVIC_Overshoot}} \label{appendix:LfDVIC_LMI_Proof_Overshoot}

\begin{figure}[!tbp]
	\centerline{\includegraphics[width=0.75\linewidth]{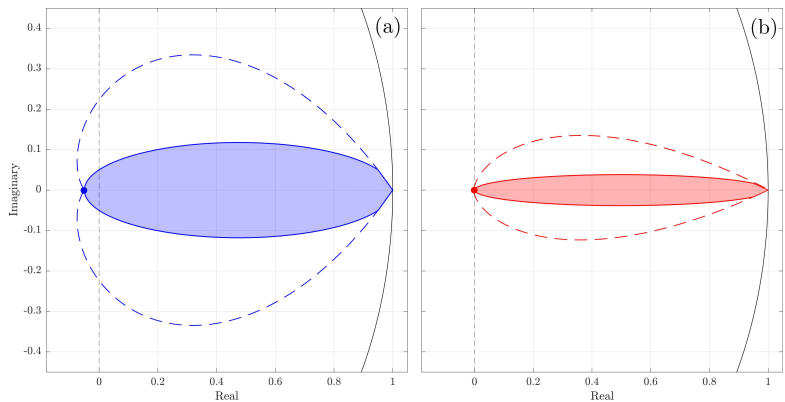}}
	\caption{Representation of the $\mathbb{D}$-stability region (shaded) approximating the non-convex cardioid defined by the logarithmic spirals (dotted-lines) region in the complex plane generated for maximum percentage overshoots of $5\%$ (a) and $0.01\%$ (b).}
	\label{fig:D_stab}
	%\vspace{-5mm}
\end{figure}

Condition~\eqref{eq:LfDVIC_LMI_Dstab_OS} corresponds to the $\mathbb{D}$-stability definition~\cite{peaucelle2000new}, which is a generalisation of the Lyapunov stability to limit the eigenvalues of state matrix $\StateMatDisc(k)$ to a region $\mathbb{D}$ in the complex plane symmetric w.r.t. real axis and defined through \LDstab and \MDstab. A maximum percentage overshoot w.r.t. a reference value $\overline{\PercOverShoot}$ in second order systems corresponds to a maximum damping ratio \DampingRatio as defined in~(\ref{eq:LfDVIC_Dstab_OS_def_terms}a). This corresponds in the (discrete) complex plane to a logarithmic spiral, characterised by \LogSpiral, which determines its intersection with the real axis \IntersectOS as defined in~Eqs.~(\ref{eq:LfDVIC_Dstab_OS_def_terms}b,c). In~\cite{Rosinova2019}, the non-convex region (namely cardioid) generated by these spirals is approximated by the intersection of two regions: an ellipsoid and a cone, defined in Eqs.~(\ref{eq:LfDVIC_Dstab_OS_def_terms}a,b) and~(\ref{eq:LfDVIC_Dstab_OS_def_terms}c,d) respectively. The ellipsoid is defined by its center \CenterOS~(Eq.~(\ref{eq:LfDVIC_Dstab_OS_def_terms}d)) and its major and minor axes \MajorAxOS~(Eq.~(\ref{eq:LfDVIC_Dstab_OS_def_terms}e)) and \MinorAxOS~(Eq.~(\ref{eq:LfDVIC_Dstab_OS_def_terms}f)); and the cone with its vertex in $(1,0)$ by half its inner angle \ConeAngle~(Eq.~(\ref{eq:LfDVIC_Dstab_OS_def_terms}g)). To define both regions, an intersection point between  them $r=(a,b)$must be specified such that it belongs to the logarithmic spiral. In this paper, a value of $a=0.95$ has been chosen as it has been seen to provide a good approximation closer to $(0,1)$. Figure~\ref{fig:D_stab} depicts the cardioid regions and their approximations for different $\overline{\PercOverShoot}$ values. \hfill $\blacksquare$

%-------------------------------------------------------------------------------------------------------------------------------------------------------------------------------------------------------------------------------

\vspace{5mm}

\section{Implementation details}\label{appendix:LfDVIC_method_impl}

Controller design process has been carried out within \textsc{MATLAB} environment using 
R2109b version on an Intel Core i7-9700K CPU @3.70GHz×8 with a NVIDIA GP106 GPU. 

\vspace{2mm}
\subsection{H-GP model generation}\label{appendix:LfDVIC_HGPGen}

Using the available GP implementation in \textsc{MATLAB}, we consider for the H-GP model generation as convergence criterion an improvement between iterations below $5\%$ w.r.t the difference between the two noise prediction means. Thus, for the validation experiment, using a set of $10$ taught position trajectories consisting on 240 samples each (i.e. a sampling time of $0.02$[s]), the method takes (on average) $9$ iterations to converge, which corresponds to an average time of $45.136$[s] (for each DoF).

\vspace{2mm}
\subsection{Controller solution search} \label{appendix:LfDVIC_SolSearch}

The \textit{off-the-shelf} Bayesian Optimisation Algorithm from \textsc{MATLAB} has been used as the solution-search method. For convergence criterion, we have considered $\epsilon = 0.01$ (below $1\%$ for the fitness range $[0,1]$) and $N_{\text{iter}} = 75$. The LMI problem has been formulated through the \textsc{YALMIP} toolbox~\footnote[1]{\textsc{YALMIP} toolbox : \url{https://yalmip.github.io/}} (release $2018-10-12$), and solved with the semi-definite programming algorithms provided by \textsc{MOSEK}~\footnote[2]{\textsc{MOSEK} softw: \url{https://www.mosek.com}}(version 9.3.10). Thus (on average) each iteration takes $1.737$[s] from which the generation of the LMI constraints $0.028$[s] ($1.6\%$), the execution through \textsc{YALMIP} $0.0570$[s] ($3.28\%$) and \textsc{MOSEK} solver $0.005$[s] ($0.3\%$). This means that a problem setting that requires $250$ iterations for convergence will take approx $7.3$[min]. The average number of iterations until convergence for all the User Preferences and Designs used in the validation are included in Table~\ref{tab:mean_iterations}. 

\begin{table}[!t]
	\footnotesize
	\centering
	\begin{tabular}{|c|c|c|c|c|}
		\hline
		\diagbox[width=7em]{\textbf{Design}}{\textbf{User}\\\textbf{Pref.}}  & \textbf{I} & \textbf{II} & \textbf{III} & \textbf{IV} \\ \hline
		\textbf{A} & 167  &  170  &  263  &  419 \\ \hline
		\textbf{B} & 229  &  208  &  322  &  576 \\ \hline
		\textbf{C} & 156  &  157  &  272  &  504 \\ \hline
		\textbf{D} & 180  &  176  &  415  &  630 \\ \hline
	\end{tabular}
	\caption{Number of iterations until convergence for each User Preference - LMI Design combination.}
	\label{tab:mean_iterations}
\end{table}

\end{appendices}

%-----------------------------------------------------------------------------------------------------------------------------------------------------------------------------------------------

%% appendix sections  then done as normal sections



\bibliographystyle{IEEEtran}
\bibliography{library.bib}

\end{document}